\documentclass{article}
\usepackage[margin=0.7in]{geometry}
\usepackage{graphicx} 
\usepackage[nottoc]{tocbibind}
\usepackage{color}
\usepackage{multirow}
\usepackage{chngpage}
\usepackage{amsmath}
\usepackage{makecell}
\usepackage{algorithm,algorithmic}

\usepackage[nolist,nohyperlinks]{acronym}
\usepackage{multirow}
\usepackage{graphicx} 
\usepackage[nottoc]{tocbibind}
\usepackage{xcolor}
\usepackage{soul}
\usepackage{multirow}
\usepackage[colorlinks=true, urlcolor=blue]{hyperref}
\usepackage{chngpage}
\usepackage{amsmath}
\usepackage{makecell}
\usepackage{algorithm,algorithmic}
\sethlcolor{yellow}
\usepackage[switch]{lineno}
\usepackage{bbm}
\usepackage{authblk}
\usepackage{amssymb}

\usepackage[nolist,nohyperlinks]{acronym}
\acrodef{AI}{Artifical Intelligence}
\acrodef{FL}{Federated Learning}
\acrodef{NLP}{Natural Language Processing}
\acrodef{M2D}{Model to Data}
\acrodef{D2M}{Data to Model}
\acrodef{M2M}{Model to Model}
\acrodef{ANN}{Artificial Neural Network}
\acrodef{HMM}{Hidden Markov Model}
\acrodef{SVM}{Support Vector Machine}
\acrodef{DT}{Decision Tree}
\acrodef{GAN}{Generative Adversarial Network}
\acrodef{ML}{Machine Learning}
\acrodef{DL}{Deep Learning}
\acrodef{RL}{Reinforcement Learning}
\acrodef{GRNN}{Generative Regression Neural Network}
\acrodef{P3}{Privacy Preserving Photo Sharing}
\acrodef{GMI}{Generative Model-Inversion}
\acrodef{DLG}{Deep Leakage from Gradients}
\acrodef{iDLG}{Improved DLG}
\acrodef{IG}{Inverting Gradient}
\acrodef{MSE}{Mean Square Error}
\acrodef{CD}{Cosine Distance}
\acrodef{PSNR}{Peak Signal-to-Noise Ratio}
\acrodef{WD}{Wasserstein Distance}
\acrodef{TVLoss}{Total Variation Loss}
\acrodef{BN}{Batch Normalization}
\acrodef{GGL}{Generative Gradient Leakage}
\acrodef{CMA-ES}{Covariance Matrix Adaptation Evolution Strategy}
\acrodef{BO}{Bayesian Optimization}
\acrodef{DP}{Differential Privacy}
\acrodef{HE}{Homomorphic Encryption}
\acrodef{MPC}{Secure Multi-Party Computation}
\acrodef{PRECODE}{Privacy Enhancing Module}
\acrodef{IID}{Independent and Identically Distributed}
\acrodef{non-IID}{non-Independent and Identically Distributed}
\acrodef{PGD}{Projected Gradient Descent}
\acrodef{APA}{Accumulative Poisoning Attack}
\acrodef{FedRec}{Federated Recommendation}
\acrodef{ASR}{Attack Success Rate}
\acrodef{MarMed}{Marginal Median}
\acrodef{MeaMed}{Mean-around-Median}
\acrodef{DBA}{Distributed Backdoor Attack}
\acrodef{F3BA}{Focused Flip Federated Backdoor Attack}
\acrodef{DDif}{Division Differences}
\acrodef{NEUP}{NormalizEd UPdate Energy}
\acrodef{PCA}{Principal Component Analysis}
\acrodef{NAD}{Neural Attention Distillation}

\title{A Survey on Vulnerability of Federated Learning: A Learning Algorithm Perspective}
\author[1]{Xianghua Xie\thanks{Corresponding Author: \href{mailto:x.xie@swansea.ac.uk}{x.xie@swansea.ac.uk}}}
\author[1]{Chen Hu}
\author[1]{Hanchi Ren} 
\author[2]{Jingjing Deng\thanks{Corresponding Author: \href{mailto:jingjing.deng@durham.ac.uk}{jingjing.deng@durham.ac.uk}}}
\affil[1]{Department of Computer Science, Swansea University, United Kingdom}
\affil[2]{Department of Computer Science, Durham University, United Kingdom}    
\date{}                     
\begin{document}

\maketitle

\begin{abstract}
Federated Learning (FL) has emerged as a powerful paradigm for training Machine Learning (ML), particularly Deep Learning (DL) models on multiple devices or servers while maintaining data localized at owners' sites. Without centralizing data, FL holds promise for scenarios where data integrity, privacy and security and are critical. However, this decentralized training process also opens up new avenues for opponents to launch unique attacks, where it has been becoming an urgent need to understand the vulnerabilities and corresponding defense mechanisms from a learning algorithm perspective. This review paper takes a comprehensive look at malicious attacks against FL, categorizing them from new perspectives on attack origins and targets, and providing insights into their methodology and impact. In this survey, we focus on threat models targeting the learning process of FL systems. Based on the source and target of the attack, we categorize existing threat models into four types, \ac{D2M}, \ac{M2D}, \ac{M2M} and composite attacks. For each attack type, we discuss the defense strategies proposed, highlighting their effectiveness, assumptions and potential areas for improvement. Defense strategies have evolved from using a singular metric to excluding malicious clients, to employing a multifaceted approach examining client models at various phases. In this survey paper, our research indicates that the to-learn data, the learning gradients, and the learned model at different stages all can be manipulated to initiate malicious attacks that range from undermining model performance, reconstructing private local data, and to inserting backdoors. We have also seen these threat are becoming more insidious. While earlier studies typically amplified malicious gradients, recent endeavors subtly alter the least significant weights in local models to bypass defense measures. This literature review provides a holistic understanding of the current FL threat landscape and highlights the importance of developing robust, efficient, and privacy-preserving defenses to ensure the safe and trusted adoption of FL in real-world applications. The categorized bibliography can be found at: \url{https://github.com/Rand2AI/Awesome-Vulnerability-of-Federated-Learning}.
\end{abstract}

\section{Introduction}
\label{sec:introduction}

In the era of \ac{AI} that is built upon big data, the need to extract valuable insights from massive amounts of information is driving innovation across industries. Achievements of data-driven \ac{DL} models have been witnessed in many areas, ranging from \ac{NLP}~\cite{radford2018improving,radford2019language,brown2020language} to visual computing~\cite{ho2020denoising,sohl2015deep,song2019generative,song2020score}. It is generally agreed upon that the more training data, the greater potential performance of the model. To illustrate, the research work~\cite{nature-review} claims if one were able to collect data from all medical facilities, models trained on such dataset would have the potential of ''answering many significant questions'', such as drug discovery and predictive modeling of diseases. Data centralization scheme for training \ac{AI} model has been the predominant method for decades.
However, methods solely relying on centralized training scheme are becoming less viable, not only due to the cost of computational resources, but more importantly, the growing concerns related to privacy and security,
which has triggered the need for alternative learning paradigms. \ac{FL}~\cite{konevcny-communicate,google-fl}, a distributed learning paradigm emerges as a pioneering solution to address these challenges, where multiple decentralized parties collaborate on a learning task while the data remains with its owner. In contrast to traditional approaches, where all data has to be centralized, \ac{FL} stemming from the increasing concerns on data privacy allows model to be trained at the source of data creation. This innovative approach not only minimizes the risk of data leakage, maintains the privacy of sensitive information, but also lifts the computational burden of cloud centers, which is considered as a potential alternative for completing multi-party learning in many domains, such as: healthcare~\cite{antunes2022federated,nguyen2022federated,xu2021federated}, finance~\cite{long2020federated,byrd2020differentially,yang2019ffd}, smart cities~\cite{zheng2022applications,jiang2020federated,nguyen2021federated} and autonomous driving~\cite{zhang2021end,nguyen2022deep,zhang2021real}. We observed that there is a significant growth related to \ac{FL} in both academic research and industrial applications.

\begin{figure*}[ht!]
\centering
\includegraphics[width=0.9\linewidth]{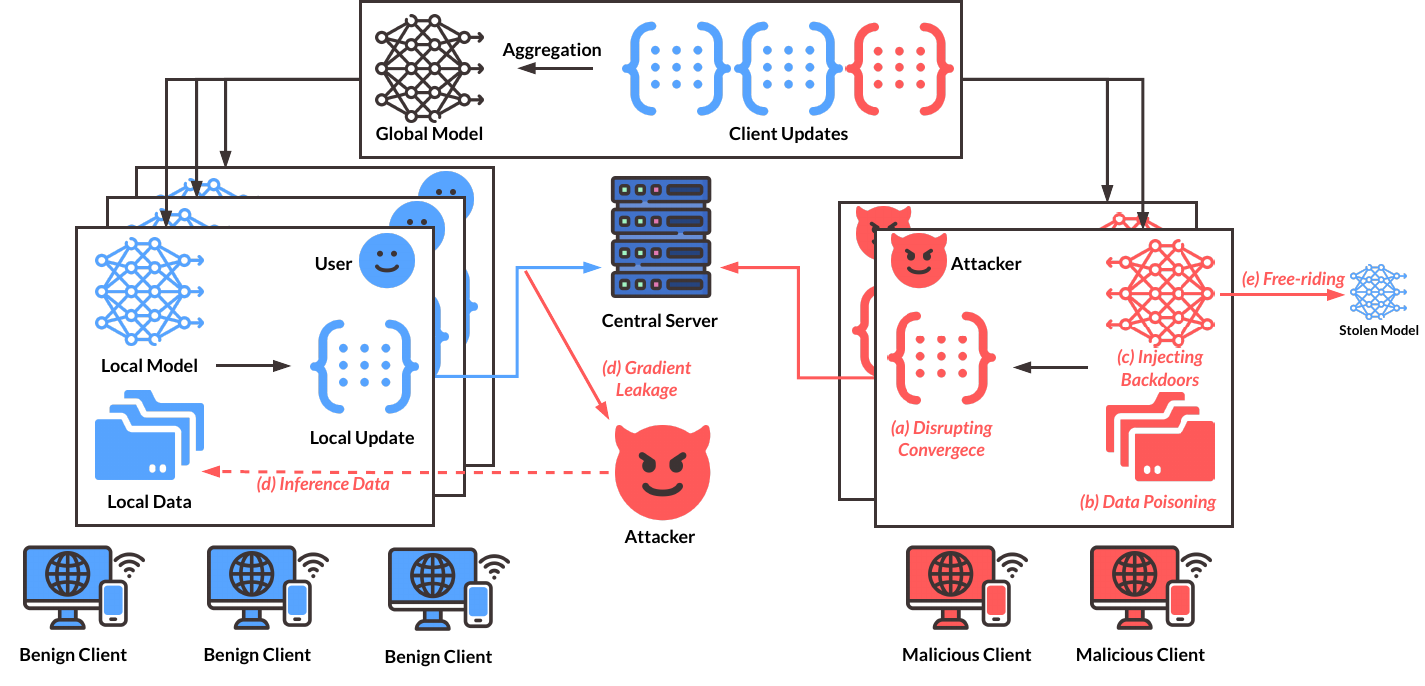}
\caption{An Overview of Common Vulnerabilities in \ac{FL}. Malicious attackers can: (a) manipulate model updates to prevent the global model from converging; (b) tamper data labels to induce erroneous predictions after training; (c) inject backdoors into the global model; (d) reconstruct data or inference data properties by eavesdropping model updates; (e) steal the global model while contribute nothing.}
\label{fig: overview}
\end{figure*}

Recent studies on exploiting vulnerabilities of \ac{FL}, have illuminated the fact that the robustness of \ac{FL} architectures is not as secure as expected, where each building block in \ac{FL} algorithms, ranging from its data distribution, communication mechanisms, to aggregation processes, is susceptible to malicious attacks~\cite{open-problems,threats-survey,grad_inver_survey,vfl-survey}. These vulnerabilities can potentially compromise the privacy and security of the participants, meanwhile downgrade the integrity and effectiveness of the entire learning system. Figure~\ref{fig: overview} illustrates various common \ac{FL} attacks and provides a comprehensive overview on different stages and components in the \ac{FL} that can be targeted by opponents. Specifically, a variety of tactics that a malicious attacker can employ, as follows:
\begin{itemize}
    \item \textbf{Data Tampering:} By disrupting data label or introducing sample noisy the adversary misguides the global model making inaccurate or biased predictions.

    \item \textbf{Model Manipulation:} By changing the model weight during aggregation, the attacker forces the global model to deviate from the desirable convergence. It can be a subtle change over time, or a drastic disruption that leads to significant performance degradation.

    \item \textbf{Data Reconstruction:} By exploring the gradient information or model weight, the opponent attempts to reconstruct or infer specific attributes of the original data, thereby breaching the privacy of data owner.

    \item \textbf{Backdoor Injection:} By embedding backdoor into the global model, the contestant deceives the trained model to give designated prediction when the corresponding trigger pattern in the input is presented.
\end{itemize}

\begin{table*}[ht!]
\setlength{\tabcolsep}{5pt}
\renewcommand{\arraystretch}{1.5}
\centering
\caption{Our proposed taxonomy}
\begin{tabular}{ c c c } 
\hline
\textbf{Type of Attack} & \textbf{Definition} & \textbf{Example} \\
\hline
\hline
\makecell[c]{Data to Model (D2M)} & \makecell[c]{tampering the data alone to degrade model performance} & label-flipping \\ 
\hline
\makecell[c]{Model to Model (M2M)} & \makecell[c]{tampering updates to prevent learning convergence} & Byzantine attack \\
\hline
\makecell[c]{Model to Data (M2D)} & \makecell[c]{intercepting model updates to inference private data information} & gradient leakage \\
\hline
Composite (D2M+M2M) & \makecell[c]{tampering both data and updates to manipulate model behavior} & backdoor injection\\
\hline
\end{tabular}
\label{tab:taxonomy}
\end{table*}

Despite the promising future of \ac{FL} aimed at alleviating privacy concerns, \ac{FL} still faces a wide variety of threats. In contrast to reviewing \ac{FL} from system and network security perspectives, in this survey, we focus on retrospecting the research advancements of \ac{FL} vulnerability that is inherited from the nature of machine learning algorithms. As shown in Figure~\ref{fig: overview}, we identify that a malicious attacker can attack every component in the \ac{FL} system. For example, an opponent may masquerade as a participating client of the system and provide toxic data to degrade the prediction performance of the global model, or intercept client updates and inject backdoor or reconstruct private training data. 
In this paper, we propose a taxonomy of \ac{FL} attacks centered around attack origins and attack targets, which are outlined in Table~\ref{tab:taxonomy}. Our taxonomy of \ac{FL} attacks emphasizes exploited vulnerabilities and their direct victims. For instance, label-flipping is a typical \ac{D2M} attack, often described as a data poisoning technique. If the local data is tampered by such a designated attack, the trained global model can be compromised by such training data and exhibit anomalous behavior.

The rest of survey is organized as such: In Section 2, we firstly introduce the essential preliminaries of \ac{FL} algorithm. Then, following the proposed taxonomy, we review each type of attack, including \ac{D2M} Attack, \ac{M2M} Attack, \ac{M2D} Attack and Composite Attack in Section 3, 4, 5 and 6 respectively. Within each section, both threat models and the corresponding defense strategies are presented, compared and discussed. Section 7 concludes our findings and provides our recommendation for future research directions.

\section{Preliminaries of Federated Learning}

\ac{FL} can be categorized into horizontal FL, vertical FL, and federated transfer learning, based on how the training data is organized \cite{non-iid-review}. Since the majority of research on FL vulnerabilities focuses on the horizontal FL setting, therefore, we also focus on horizontal FL as the central topic in this review. FedAvg is the most classic horizontal \ac{FL} algorithm, where the global model is learned by averaging across all local models trained on clients. Surprisingly, such a simple aggregation scheme has been proven to be effective in many case studies \cite{fedgan, fednlp, fedod}, where the convergence is also mathematically sound~\cite{fed-convergence}. Improvements upon FedAvg include incorporating local update corrections \cite{scaffold, FedProx} or adaptive weighting schemes \cite{fedatt, fedmed, ren2020fedboost}, however, the fundamental aggregation scheme remains similar. Therefore, we present FedAvg~\cite{google-fl} as an example to demonstrate the potential components in FL system that can be targeted by malicious parties. Firstly, all clients receive the identical global model $\omega_0$ from the central server that is randomly initialized. Then, the local model is trained on each client with its local data. Once the local training steps finish (i.e., the number of pre-set iteration or epoch is reached), individual clients send either the updated local model $\omega_{E}$ or the model difference $u$ to the server. The central server aggregates the global model $\omega_{r}$ by averaging the local models, and send the updated model to each client. To speed up the training, a subset of clients are chosen randomly for the current round of training, which is also considered as a dropout regularization for FL. The pseudo code of original FedAvg algorithm is given in Algorithm~\ref{alg:fedavg}, where the terms highlighted indicate the entities that can be compromised. 

The comparison between surveys on \ac{FL} attacks and defenses is summarized in Table \ref{tab:review-comparison}. While most surveys include detailed discussion on defense strategies, some of them only give high-level overviews on threat models, such as explaining the concept of Byzantine attacks (M2M) without delving into diverse attacks as we summarized in Table \ref{tab:M2M}. Our work reviews \ac{FL} vulnerabilities from the perspective of learning algorithms. Our review includes major threat models that exploits the learning paradigm of \ac{FL} and discusses defense strategies to counter these threats.

\begin{table*}[ht!]
\setlength{\tabcolsep}{5pt}
\centering
\caption{Comparison of Related Surveys on Federated Learning Attacks and Defenses}
\begin{tabular}{ c | c c c c c c c c } 
\hline
\multirow{3}{*}{Surveys} & \multicolumn{8}{c}{Federated Learning Attacks and Defenses} \\
\cline{2-9}
 & \multicolumn{2}{c|}{D2M} & \multicolumn{2}{c|}{M2M} & \multicolumn{2}{c|}{M2D} & \multicolumn{2}{c}{Composite} \\
\cline{2-9}
  & \multicolumn{1}{c|}{Threat} & \multicolumn{1}{c|}{Defense} & \multicolumn{1}{c|}{Threat} & \multicolumn{1}{c|}{Defense} & \multicolumn{1}{c|}{Threat} & \multicolumn{1}{c|}{Defense} & \multicolumn{1}{c|}{Threat} & \multicolumn{1}{c}{Defense} \\
\hline
Kairouz et al. \cite{open-problems} & $\circ$ & $\checkmark$ & $\circ$ & $\checkmark$ & & & $\circ$ & $\checkmark$ \\
Nguyen et al. \cite{new-backdoor-review} & $\checkmark$ & $\checkmark$ & $\checkmark$ & $\checkmark$ & & & $\checkmark$ & $\checkmark$ \\
Zhang et al. \cite{trustworthy-review} & $\circ$ & $\checkmark$ & $\circ$ & $\checkmark$ & $\circ$ & $\checkmark$ & $\circ$ & $\checkmark$ \\
Gong et al. \cite{brief-backdoor-review} & $\circ$ & & $\circ$ & & & & $\circ$ & $\checkmark$ \\
Yin et al. \cite{privacy-preserving-review} & & & & & $\checkmark$ & $\checkmark$ & & \\
Zhang et al. \cite{security-threats-review} & $\circ$ & $\checkmark$ & $\circ$ & $\checkmark$ & $\circ$ & $\checkmark$ & & \\
\hline
ours & $\checkmark$ & $\checkmark$ & $\checkmark$ & $\checkmark$ & $\checkmark$ & $\checkmark$ & $\checkmark$ & $\checkmark$ \\
\hline
\multicolumn{4}{l}{\small $\circ$: high-level overview \ \ $\checkmark$: detailed review.}
\end{tabular}
\label{tab:review-comparison}
\end{table*}

\begin{algorithm}[ht!]
    \caption{FedAvg for Horizontal FL. (\hl{$\textcolor{red}{Terms}$} highlighted are the vulnerable components can be targeted by adversaries.)}
    
    $n_i$ is the number of local samples, $N_S$ is the total number of samples among selected clients, $D_i$ is the local training data, $\omega$ is model weights
    
    \textbf{Server:}
    \begin{algorithmic}[1]
        \STATE create and send model to all clients
        \STATE clients own their respective data $D_i$
        \STATE initialize $\omega_{0}$
        \FOR{each round $r = 1,2,...,R$}
            \STATE sample $|S|$ clients, send $\omega_{r-1}$ to each clients in $S$
            \FOR{each client $i \in S$}
                \STATE $\omega_{r}^i$ or $u_i \leftarrow $Client$(i,$ \hl{$\textcolor{red}{\omega_{r-1}}$})
            \ENDFOR
            \STATE $\omega_{r} \leftarrow \sum_{i=1}^{|S|}$ \hl{$\frac{\textcolor{red}{n_i}}{N_S}$} $\omega_{r}^i$ or $\omega_r \leftarrow \omega_{r-1} + \sum_{i=1}^{|S|}$ \hl{$\frac{\textcolor{red}{n_i}}{N_S}$} $u_i$
            \STATE validate the model with $\omega_{r}$
        \ENDFOR
    \end{algorithmic}
    \textbf{Client($i, \omega$):}
    \begin{algorithmic}[1]
        \FOR{each epoch $e = 1,2,...,E$}
            \STATE $\omega_{e} \leftarrow$ \hl{$\textcolor{red}{\omega_{e-1}}$} $- \eta \cdot$ \hl{$\textcolor{red}{\nabla_{\omega_{e-1}}}$} $\mathcal{L}($\hl{$\textcolor{red}{D_{i}}$}$)$
        \ENDFOR
        \STATE \hl{$\textcolor{red}{u}$} $ \leftarrow \omega_{E} - \omega$
        \STATE return $\omega_{E}$ or $u$ to server
    \end{algorithmic}
    \label{alg:fedavg}
\end{algorithm}

\section{Data to Model Attacks}

\begin{figure*}[ht!]
\centering
\includegraphics[width=0.7\linewidth]{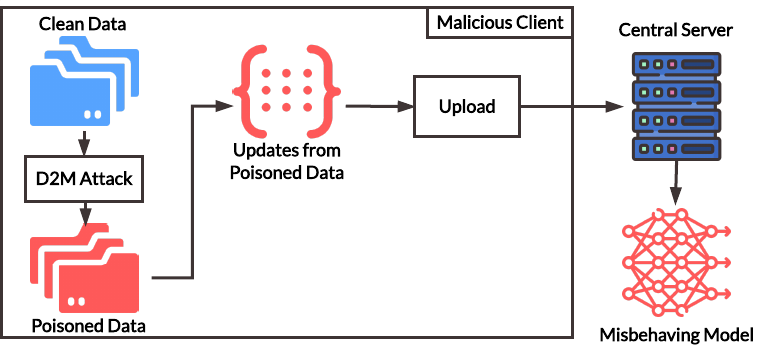}
\caption{An illustration for D2M attack.}
\label{fig: D2M attack}
\end{figure*}

\begin{figure*}[ht!]
\centering
\includegraphics[width=1.0\linewidth]{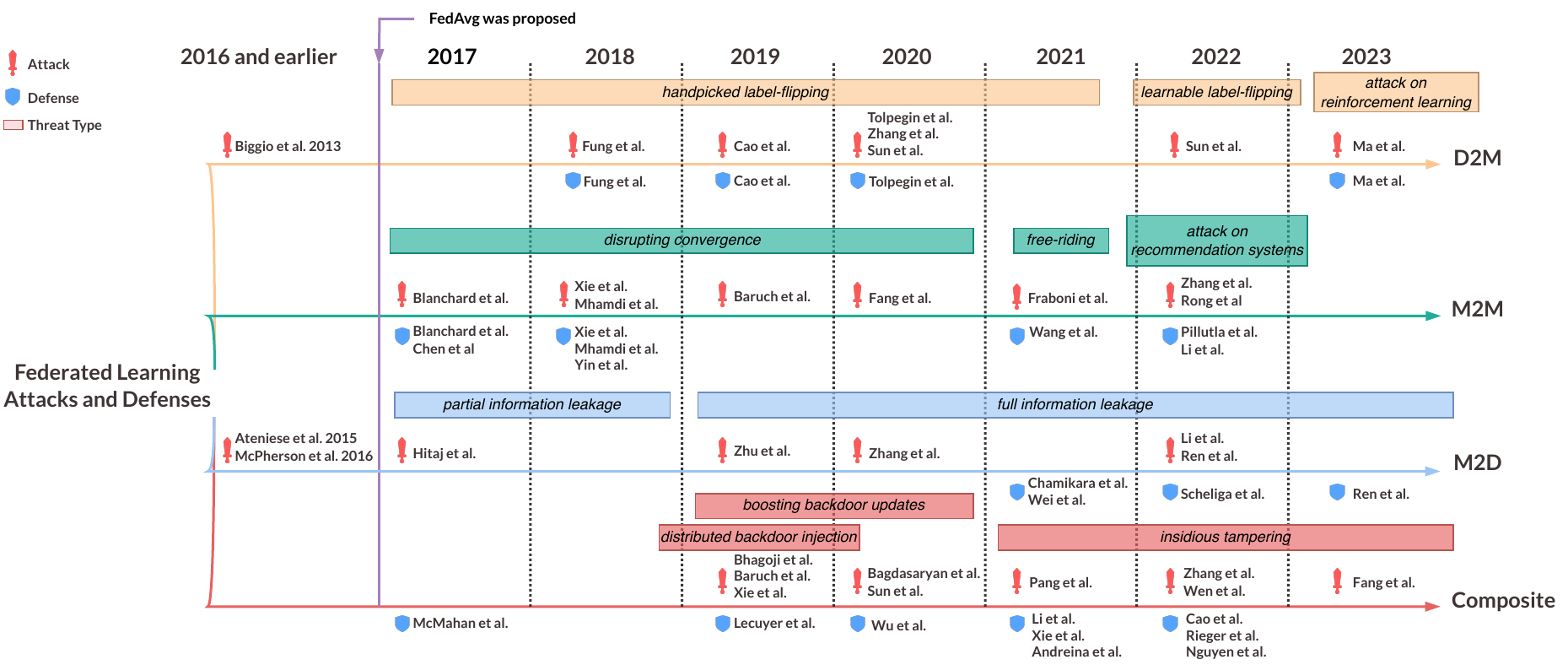}
\caption{The timeline of research on \ac{FL} attacks and defenses.}
\label{fig: D2M aLine}
\end{figure*}

We describe Data to Model (D2M) attacks in \ac{FL} as threat models that are launched by manipulating the local data while the models in training are being targeted as victims. \ac{D2M} attacks are also considered as black-box attacks because the attackers do not need to access inside information such as client model weights or updates, tampering the data alone is often suffice to launch a \ac{D2M} attack. However, the attackers can also draw information from local dataset or client models to enhance the effectiveness of \ac{D2M} attacks. We present the timeline of \ac{D2M} research in Figure~\ref{fig: D2M aLine}. The characteristics of discussed \ac{D2M} attacks are shown in Table~\ref{tab:d2m}.

\subsection{D2M Attacks on Class Labels} 

The \ac{D2M} attack of poisoning data labels is called label-flipping. Such an attack aims at misleading the training models by feeding tampered labels for training. For instance, the attackers may switch the labels for car images to ``planes'', resulting in the model to classify car images as planes after training.

Label-flipping attack is first studied and proved its effectiveness in the centralized setting \cite{svm-labelflip}. Later on, \cite{sybil-attack, detailed-label-flip} demonstrate label-flipping attack in \ac{FL} scenarios. Theses studies follow \cite{svm-labelflip} and flip the labels from the victim class to a different target class. Authors of \cite{detailed-label-flip} show that with only 4\% of total clients being malicious, label-flipping attack can cause the recall on victim class to drop by 10\% on the Fashion-MNIST dataset \cite{fashion-mnist}, indicating that even a small number of malicious clients can effectively degrade the performance of a defenseless \ac{FL} system through label-flipping attack. In PoisonGAN \cite{poisonGAN}, the label-flipping attack is further improved. Targeting a \ac{FL} system for image classification, the authors of PoisonGAN use the global model received on clients as the discriminator for \ac{GAN}. The attacker trains a local generator until the global model classifies generated images as the victim class. The attackers can then flip labels of generated images, compromising client models by feeding fake images along with flipped labels. The noteworthy advantage of PoisonGAN is that the attacker now does not need to access clients' data. The attacker can simply generate their own poisonous data samples. Instead of arbitrarily choosing the target class to flip, studies such as \cite{yes-we-can, semi-target} investigate different heuristic for choosing the target class. Semi-targeted attack proposed in \cite{semi-target} uses distance measures to determine which target class can more easily affect model predictions. The intuition of this attack is that if samples of two different classes are relatively close in the feature space, then label-flipping attack on these two classes is more likely to succeed as the proximity of features suggests easier learning convergence. The authors of \cite{semi-target} consider both the \ac{IID} and \ac{non-IID} scenarios. If client data is \ac{IID}, the attacker uses the global model to extract features for the local training data. The geometric center of each class is computed based on features of local data and the target class should be the one closest to the victim class. In the \ac{non-IID} scenario, the local feature space no longer well represents the structure of the global feature space. Thus, the authors leverages the scale of updates to measure which class is closer to the victim class. The attacker feeds local samples of the victim class to the global model and examines the scale of gradients when these samples are annotated as different classes. The class label that induces the smallest gradient is chosen as the target class. Different from \cite{poisonGAN, semi-target} that exploit the global model for their attacks, the heuristic of the edge-case attack \cite{yes-we-can} is built on the distribution of the training data. The edge-case attack flips labels into classes in the tail of the data distribution. Although the edge-case attack only affects a minority of samples, it can severely impair the model's fairness for underrepresented input and may pose great threats in autonomous driving systems \cite{yes-we-can}. Experiments in \cite{yes-we-can} show that the attack is most effective when the attacker holds most of the edge samples. As honest clients possess larger portions of edge samples, the attack is erased by benign updates.

\begin{table*}[ht!]
\setlength{\tabcolsep}{18pt}
\renewcommand{\arraystretch}{1.5}
\centering
\caption{Characteristics of \ac{D2M} Attacks.}
\begin{tabular}{ c c c } 
\hline
\textbf{Threat Model} & \textbf{Threat Objective} & \textbf{Poisoned Data} \\
\hline
\hline
\makecell[c]{Label-Flipping \cite{sybil-attack, detailed-label-flip, understanding-label-flipping}} & mislassification & class labels  \\
\hline
\makecell[c]{Semi-Target Poisoning \cite{semi-target}} & misclassification & class labels \\
\hline
\makecell[c]{Edge-case Attack \cite{yes-we-can}} & misclassification & class labels \\
\hline
AT$^2$FT \cite{bilevel-data-opt} & misclassification & general samples \\
\hline
PoisonGAN \cite{poisonGAN} & misclassification & \makecell[c]{general samples and\\ class labels} \\
\hline
\makecell[c]{Covert Channel \cite{secret-channel}} & \makecell[c]{secretly passing messages} & edge samples \\
\hline
\makecell[c]{Fake Sample Size \cite{fake-size-review}} & disrupting convergence & client dataset size \\
\hline
\makecell[c]{Local Environment Poisoning \cite{frl-d2m}} & poisoning policy & agent rewards \\
\hline
\makecell[c]{Poisonous Ratings \cite{frs-doublepack}} & \makecell[c]{controlling item\\ recommendation} & item ratings \\
\hline
\end{tabular}
\label{tab:d2m}
\end{table*}

\subsection{D2M Attacks on Samples}

Labels are not the only target in \ac{D2M} attacks. Depending on the \ac{FL} scenario, the attackers may choose to poison other relevant client data. A threat model that targets the sample size on clients is proposed in \cite{fake-size-review}. Based on the fact that FedAvg computes the weighted average of client weights based on the numbers of their corresponding local samples, the attacker can simply falsely report the number of local samples to be a large number such that the aggregated model will be dominated by the attacker's chosen model. AT$^2$FT \cite{bilevel-data-opt} is another \ac{D2M} attack that generate poisonous samples. The difference between AT$^2$FT and PoisonGAN \cite{poisonGAN} is that the former does not flip labels. Authors of AT$^2$FT formulates their attack as a bilevel optimization problem in which the attacker tries to perturb subsets of local training samples such that losses on local clean data are maximized. In essence, the AT$^2$FT algorithm maximizes local losses through gradient ascent where gradients \emph{w.r.t} the perturbed data are approximated by minimizing a dual problem. The \ac{D2M} attacks are also not limited to classification tasks. The authors of \cite{frl-d2m} propose a \ac{D2M} threat model, local environment poisoning, targeting federated \ac{RL}. The attacker can influence the learned policy by providing fake rewards during local agent training. Fake rewards are derived from gradient descent such that they minimize the objective function of \ac{RL}. A \ac{D2M} threat model on \ac{FedRec} systems is proposed in \cite{frs-doublepack}. Specifically, the authors of \cite{frs-doublepack} focused on the graph neural network based \ac{FedRec} system proposed in \cite{frs-gnn}. By feeding compromised client models with fake item ratings during training, the attacker can force the recommendation system to show specified item ratings for specific users.

Unlike the above methods that use \ac{D2M} attacks to influence model predictions, the covert channel attack proposed in \cite{secret-channel} aims at secretly transmitting messages between two clients. On the receiver client, the attacker first looks for edge samples from its local training data such that even a small perturbation in the data results in different classification outcomes. Perturbed edge samples along with the transmission interval, the clean and poisoned class predictions are sent to the sender client. The sender client decides whether to fine-tune its local model with the perturbed data depending on the message bit it wishes to send and the local model's prediction. Once the receiver client receives the updated model, it can decode the message bit based on the classification outcome of perturbed samples.

For \ac{D2M} attacks to be successful, studies in \cite{sybil-attack, detailed-label-flip, understanding-label-flipping} show that it is vital to ensure the availability of malicious clients during training. If no malicious client are selected to participate in the global model update, the effects of their attacks can be quickly erased by updates from benign clients \cite{detailed-label-flip}. Recent studies on \ac{FL} threat models tend to combine \ac{D2M} attacks with \ac{M2M} attacks to launch more powerful composite attacks. Since the attacker also manipulates model updates, composite attacks can be stealthier and more persistent. Such attacks also give the attacker more freedom of when and how to trigger the attack.

\subsection{Defense Against D2M Attacks}

In this section we introduce defense strategies proposed along with studies on label-flipping attacks \cite{sybil-attack, detailed-label-flip, understanding-label-flipping, frl-d2m}. Since \ac{D2M} attacks ultimately induce changes in model updates, \ac{FL} system administrators may also consider defense mechanisms designed for \ac{M2M} or composite attacks.

Strategies proposed in \cite{sybil-attack} and \cite{detailed-label-flip} are both inspired by the observation that gradients in \ac{FL} behave differently in terms of benign and malicious clients. In particular, because of the \ac{non-IID} nature of data, it is observed in \cite{sybil-attack} that gradients from benign clients are more diverse than those from malicious clients. This is because benign gradients conform to the \ac{non-IID} distribution of local data while malicious models have a shared poisoning goal. The defense strategy FoolsGold \cite{sybil-attack} thus aims at reducing the learning rate of similar model updates while maintaining the learning rate of diverse updates. To determine the similarity of model updates, the history of all model updates are stored and pair-wise cosine similarity between current and historical updates are computed. The defense strategy in \cite{detailed-label-flip} requires prior knowledge on the attack target. This method needs the user to first choose a suspect class that is believed to be poisoned. Then only model updates directly contributing to the prediction of the suspect class are collected. These model weights subsequently go through \ac{PCA} and are clustered based on their principal components. Principal components of benign and malicious clients fall in different clusters. Similar to gradients, model weights can also be used to differentiate benign and malicious clients. Sniper \cite{understanding-label-flipping} is a defense strategy based on the Euclidean distances between model weights. The central server first computes the pair-wise distances between received client models. Then the server constructs a graph based on the distances. Client models are the nodes of the graph, and if the distance between two client models are smaller than the given threshold, these two models are then linked by an edge. If the number of models in the maximum clique of the graph is larger than half of the total number of clients, models in this clique are aggregated to update the global model. Otherwise, the server increases the distance threshold and repeat the above process until a suitable clique can be found.

Parallel learning \cite{prl} is a paradigm of \ac{RL} in which multiple agents learn concurrently to solve a problem. Parallel learning not only alleviates data deficiency but also stabilizes training, as agents learn from diverse experiences. Unlike multi-agent \ac{RL}, which aims to develop competitive or cooperative strategies among clients, parallel \ac{RL} focuses on solving single-agent problems through parallel training. This objective is similar to that of conventional federated learning, in which the goal is to obtain a global model through distributed local model training. Therefore, federated reinforcement learning becomes imperative when the learning environment of \ac{RL} is privacy-sensitive. For the \ac{D2M} threat model targeting federated \ac{RL}, a corresponding defense strategy was also proposed in \cite{frl-d2m}. This method requires the central server to evaluate client agent performance to determine their credibility. Specifically, the central server tests client policies and computes their corresponding rewards. The central server aggregates client policies based on a set of weights derived from normalized rewards.

\subsection{Evaluation Metrics for Attacks and Defenses on Classification Tasks}
\label{sec:EMD2MAD}

Since the majority of studies on D2M attacks focus on image classification, the most commonly used datasets for \ac{D2M} attack evaluation are MNIST \cite{lecun1998mnist}, Fashion-MNIST \cite{fashion-mnist} and CIFAR-10 \cite{krizhevsky2009learning}. Natural language and domain-specific datasets can also be seen \cite{sybil-attack, yes-we-can, bilevel-data-opt, frs-doublepack}. \ac{ASR} is widely used to evaluate the effectiveness of an attack. Specifically, for D2M attacks targeting classification tasks, \ac{ASR} is defined as the proportion of targeted test samples being misclassified, namely, 

\begin{equation}
    ASR = \frac{\Sigma_{(x_i, y_i) \in D} \mathbbm{1}\{f(x_i)=y_t, y_t \neq y_i\}}{|D|}
\end{equation}

\noindent where $D$ is the test set for evaluation, $x_i$ is the data sample while $y_i$ is its corresponding groundtruth label, $y_t$ is the label chosen by the attacker, $f(\cdot)$ is the attacked global model, and $\mathbbm{1}\{\cdot\}$ equals to 1 if the condition inside the brackets is met. ASR is also used to evaluate M2M or composite attacks. The metric respectively reflects how severely the attack disrupts model convergence and how sensitive the model is to backdoor triggers. In addition, the performance of the attack can also be demonstrated by the decrease in overall classification accuracy. For regression tasks, mean absolute error and root mean squared error are employed. While some defenses provide formal proof for their effectiveness, most work on \ac{FL} defenses is empirically validated by demonstrating the robustness of model performance when the defense is adopted in a malicious environment.

\section{Model to Model Attacks}

\begin{figure*}[ht!]
\centering
\includegraphics[width=0.7\linewidth]{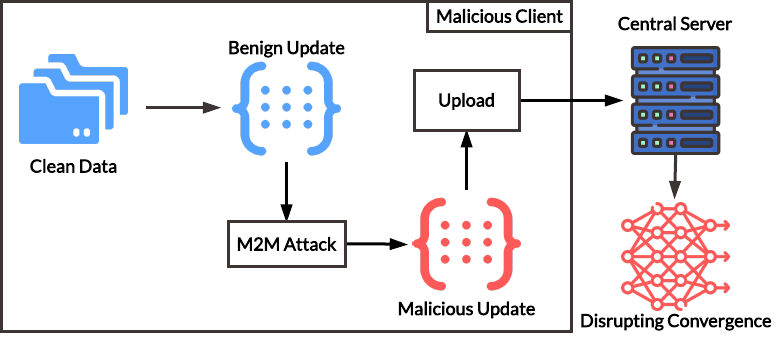}
\caption{An illustration of \ac{M2M} attack.}
\label{fig: M2M}
\end{figure*}

We define Model to Model (M2M) attacks in \ac{FL} as threat models that manipulate local model updates or weights to affect the global model, as depicted in Figure~\ref{fig: M2M}. The primary objective of an \ac{M2M} attack is to disrupt the convergence of \ac{FL} algorithms. The presence of \ac{M2M} attacks is also described as the Byzantine problem \cite{byzantine-origin}. In a distributed system affected by the Byzantine problem, benign and malicious participants coexist in the system. Malicious participants deliberately disseminate confusing or contradicting information to undermine the system's normal operations. Therefore the challenge for the system administrator lies in achieving consensus among benign participants despite the presence of malicious ones. Defending against these \ac{M2M} attacks means ensuring that the learning algorithm to converge to an optimal minima regardless of poisoned updates from malicious clients. In addition to the above threat model, a special case of \ac{M2M} attacks, called the free-rider attack, aims to steal the global model itself, infringing on the intellectual property rights of the model owner. An malicious party may pretend to join the FL system solely to obtain the distributed global model, without contributing to the learning task. Since the threat model of free-rider attack is comparatively straightforward, we discuss this type of attack along with its defense mechanisms in the same section. The characteristics of discussed \ac{M2M} attacks are shown in Table~\ref{tab:M2M}.

\begin{table*}[ht!]
\setlength{\tabcolsep}{10pt}
\centering
\caption{Various \ac{M2M} threat models}
\begin{tabular}[ht]{c|c|c|c}
\hline
\textbf{Threat Model} & \textbf{Approach} & \textbf{Type} & \textbf{Objective}  \\
\hline
free-riding \cite{free-rider} & pretend as a client & \multirow{11}{*}{a priori} & stealing global model \\
\cline{1-2}
\cline{4-4}
\makecell[c]{Byzantine Gaussian\cite{krum}} & \makecell[c]{uploading Gaussian noise} & &  \multirow{16}{*}{\makecell[c]{inhibiting convergence}}  \\
\cline{1-2}
bit-flipping \cite{generalized-bSGD} & \makecell[c]{flipping significant bits\\ of floating numbers} & & \\
\cline{1-2}
\makecell[c]{same-value attack \cite{rsa}} & \makecell[c]{uploading vectors with\\ identical values across\\ all dimensions} & & \\
\cline{1-2}
sign-flipping \cite{rsa} & \makecell[c]{flipping signs of gradients\\ on attacked clients} & & \\
\cline{1-2}
\makecell[c]{median cheating \cite{little-stat}} & \makecell[c]{cheating the aggregation\\ rule to pick the false\\ median} & & \\

\cline{1-3}
\makecell[c]{negative gradient \cite{generalized-bSGD}} & \makecell[c]{uploading the scaled\\ sum of benign gradients from\\ malicious clients} & \multirow{12}{*}{a posteriori} & \\
\cline{1-2}
norm attack \cite{buylan} & \makecell[c]{scaling certain dimensions\\ of gradients} & & \\
\cline{1-2}
\cline{4-4}
\makecell[c]{colluding attack \cite{colluding-byzantine}} & \makecell[c]{deceiving the aggregation\\ rule to pick the chosen\\ malicious client} & & \makecell[c]{converging to an inferior\\ minima} \\
\cline{1-2}
\cline{4-4}
PipAttack \cite{pipattack}& \makecell[c]{generating item embeddings\\ based on public information} & & \multirow{6}{*}{\makecell[c]{increasing $ER@K$ of target\\ items}}  \\
\cline{1-2}
FedRecAttack \cite{fedrec} & \makecell[c]{minimizing the rating\\ scores of untargeted items} & & \\
\cline{1-2}
\makecell[c]{User Approximation \cite{aha-attack}} & \makecell[c]{generating item embeddings\\ through approximated user\\ embeddings} & & \\
\hline
\end{tabular}
\label{tab:M2M}
\end{table*}

\subsection{General M2M Threat Models} 

Existing \ac{M2M} threat models can be divided into \textit{a priori} and \textit{a posteriori} attacks.\textit{A priori} attacks do not require any knowledge of benign clients while \textit{a posteriori} attacks need to forge poisonous model updates based on information from benign clients.

\subsubsection{Priori M2M Attacks}

A straightforward \textit{a priori} \ac{M2M} (prioM2M) attack is sending noise to the central server. This method is dubbed as Gaussian Byzantine in \cite{krum}. The Gaussian distribution for noise sampling often has zero mean but large variance to disrupt the convergence of the learning algorithm. Gaussian Byzantine is often used as the baseline attack \cite{generalized-bSGD, rsa}. Bit-flipping is a prioM2M attack proposed in \cite{generalized-bSGD}. On malicious clients, the bit-flipping attack flips four significant bits of certain 32-bit floating numbers in the original gradients as poisoned model updates. Another two prioM2M attacks, same-value attack and sign-flipping attack, are proposed in \cite{rsa}. For the same-value attack, malicious clients upload vectors with an identical random value on each dimension to the server. In the sign-flipping attack, malicious clients computes their own gradient as normal but flip the sign of gradients before uploading them to the central server. The prioriM2M attack proposed in \cite{little-stat} takes secure aggregation rules into account. It specifically attacks \ac{FL} systems equipped with median-based aggregation rules such as TrimMedian \cite{coordinate-median} or Krum \cite{krum}. The basic idea of the attack is to report false updates on multiple malicious clients such that with high probability the aggregation rule picks one of the malicious updates as the median for global update. The authors of \cite{little-stat} use a statistical heuristic to find the maximum deviation range which is used to forge the malicious updates. The value on each dimension of the original updates on malicious clients is transformed by the maximum deviation range to attain forged malicious updates. The authors also augment this attack with the \ac{D2M} attack, which is discussed in Section~\ref{sec:EMD2MAD}.

\subsubsection{Posteriori M2M Attacks} 

For \textit{a posteriori} \ac{M2M} (postM2M) attacks, omniscient negative gradient approach proposed in \cite{generalized-bSGD} is an equally straightforward approach compared to Gaussian Byzantine. This method assumes that the attacker have full knowledge of benign clients, then malicious clients only need to send scaled negative sum of benign gradients to the central server. The scaling factor is a large number on the order of magnitude of $10^{20}$. The postM2M attack proposed in \cite{buylan} takes Bayzantine-resilient aggregation rules into account. Specifically, this attack targets aggregation rules that compute the norms of client gradients to filter out malicious updates. The problem with norm-based aggregation rules is that $L^p$ norms cannot tell if two norms only differ in one specific dimension or every dimension. Thus, the attacker can exploit this by only poisoning one dimension of the gradients. The poisoned value can be scaled by a large factor while still being accepted by the aggregation rule as its norm is not far away from those of the benign gradients. Moreover, as the norm chosen by the aggregation rule approaches the infinite norm, the attacker can poison every dimension of model updates.

The above attacks can be launched individually on clients controlled by the attacker, these approaches does not require malicious clients to coordinate with each other. A colluding postM2M attack is later proposed in \cite{colluding-byzantine}. This method targets aggregation rules such as Krum \cite{krum} and Buylan \cite{buylan} that use the Euclidean distance between client models as the criterion for choosing trustworthy models. The threat model in \cite{colluding-byzantine} aims at pushing the global model towards the opposite of the benign update direction. To achieve this at the presence of aforementioned aggregation rules, a chosen malicious client is responsible for generating model updates that maximizes the global model update in the opposite direction. Other malicious clients generate updates that are close to the chosen one, conceiving the aggregation rules that malicious clients form a benign cluster and the chosen malicious client should be picked by the aggregation rule. 

\subsection{M2M Threat Models on Federated Recommendation Systems} 

As mentioned in the introduction section, \ac{FL} is well-suited for recommendation systems thanks to its ability to provide personalized recommendations and reduce privacy risks. A commonly used \ac{FedRec} framework is proposed in \cite{first-FR}. Research on the vulnerabilities of domain-specific \ac{FL} like \ac{FedRec} is still a nascent area. In this section, we introduce three noteworthy studies \cite{fedrec, pipattack, aha-attack} focusing on exploiting security vulnerabilities of \ac{FedRec}. 

The common goal of existing attacks on \ac{FedRec} is to increase the exposure rate of certain items. The affected recommendation system may always present or never show certain items to users. In \cite{fedrec, pipattack, aha-attack}, the attackers are assumed to only have access to item embeddings, local and global models. Embeddings that characterize users are always hidden from the attackers. In PipAttack \cite{pipattack}, the attacker increases target items' exposure rate by forging their embeddings to be similar to those of popular items. Since the attacker have no access to the popularity of items in the system, this information is retrieved from the Internet. Based on the retrieved information, the attacker locally train a popularity classifier with item embeddings as input. The weights of the classifier are then fixed, target item embeddings are poisoned by enforcing them to be classified as popular by the classifier. The poisoned item embeddings are uploaded to the central server to mislead the \ac{FedRec} system. 

Authors of FedRecAttack \cite{fedrec} later points out that major limitations of PipAttack include that it may severely degrade the recommendation performance and it needs around 10\% of clients to be attacked for it to be effective. Since the exposure rate at rank $K$ ($ER@K$) \cite{pipattack}, meaning the fraction of users whose top-$K$ recommended items include the target item, is a non-differentiable function, FedRecAttack uses a surrogate loss function to facilitate the attack. FedRecAttack also assumes that around 5\% of user-item interaction histories are publicly available for the attacker to use. The loss function of FedRecAttack encourages the rating scores of recommended non-target items to be smaller than the scores of target items with no interaction history, then the gradients of target item embeddings \emph{w.r.t} this loss function are uploaded to the central server. To further eschew being detected by secure aggregation rules, these gradients are normalized before uploading if their norms are larger than the threshold. 

Both PipAttack and FedRedAttack require public prior knowledge to work. In contrast, the $A-ra/A-hum$ attack proposed in \cite{aha-attack} does not have this requirement. $A-ra/A-hum$ also uses a surrogate loss function to promote the $ER@K$ for target items, but this attack focuses on approximating the user embeddings which are inaccessible in \ac{FedRec}. $A-ra$ assumes that the user embeddings are distributed by a zero mean Gaussian with the variance as a hyper-parameter. The attacker first samples a number of user embeddings from the Gaussian distribution, then maximized the interaction scores target items and sampled user embeddings to derive poisonous item embeddings. Instead of sampling from a Gaussian, $A-hum$ uses online hard user mining to generate user embeddings. The attacker first generate hard user embeddings that are not likely to interact with existing items. Then target item embeddings are optimized to increase their interaction chances with the synthesized hard users.

\begin{table}[ht!]
\renewcommand{\arraystretch}{1.5}
\centering
\caption{Characteristics of M2M Defenses}
\begin{tabular}{ c  c } 
\hline
\textbf{Type of Defense} & \textbf{Aggregation Criterion} \\
\hline
\hline
GeoMed\cite{geomed} & geometric median \\ 
\hline
RFA \cite{improved-gm} & \makecell[c]{Weiszfeld-smoothed \\ geometric median} \\
\hline
MarMed\cite{generalized-bSGD} & dimension-wise median \\
\hline
MeaMed\cite{generalized-bSGD} & mean-around median \\
\hline
TrimMean\cite{coordinate-median} & dimension-wise trimmed mean \\
\hline
Krum/Multi-Krum \cite{krum} & Euclidean distance \\
\hline
Bulyan\cite{buylan} & \makecell[c]{Euclidean distance and mean-\\around median} \\
\hline
ELITE\cite{elite} & gradient information gain \\
\hline
\end{tabular}
\label{tab:m2m_defense}
\end{table}

\subsection{Defense Against M2M Attack}

Because the median is robust to outliers in statistics, it is widely used in \ac{M2M} defenses to filter out malicious updates. GeoMed \cite{geomed} is an exemplar of median-based \ac{M2M} defenses. In GeoMed, the central server first divides received client gradients into multiple groups and computes the mean of each group. Then the geometric median of group means is used as the gradient for updating the global model. The approach of using geometric median for robust aggregation is further improved by authors of RFA \cite{improved-gm}. In RFA, clients compute their aggregation weights based on the aggregation rule inspired by the Weiszfeld algorithm \cite{weiszfeld}.  Including the geometric median, more median-based defenses are studied in \cite{generalized-bSGD}. \ac{MarMed} is a generalized form of median proposed in \cite{generalized-bSGD}. It computes the median on each dimension for client gradients. \ac{MeaMed} in \cite{generalized-bSGD} further leverages more values around the median. Built upon \ac{MarMed}, \ac{MeaMed} finds the top-$k$ values that are nearest to the median of each dimension, then the mean of these nearest values is used as the gradient on their corresponding dimensions. 

Besides median, trimmed mean also has the benefit of being less sensitive to outliers. The authors of \cite{coordinate-median} introduce coordinate-wise trimmed mean as an aggregation rule. For each dimension of client gradients, this rule removes the top-$k$ largest and smallest values, the mean of the remaining values is treated as the gradient on the corresponding dimension.

Another criterion for filtering out malicious updates is the Euclidean distance between norms. Krum~\cite{krum} and Bulyan~\cite{buylan} are two exemplary defenses built on this criterion. Krum is motivated by avoiding the drawbacks of square-distance or majority based aggregation rules. The problem pointed out in \cite{krum} is that malicious attackers can collude and misguide the center of norms to a bad minima for the sqaure-distance based aggregation, and the majority based aggregation is too computationally expensive as it needs to find a subset of gradients with the smallest distances among them. For a central server that adopts Krum as its aggregation rule, it first finds the $(n - f - 2)$ nearest neighbors for each client based on the Euclidean distances between their updates, where $n$ is the number of clients that participate the training, $f$ is the estimated number of malicious clients. Then the central server sums up the distances between each client and their corresponding neighbors as Krum scores. The client with lowest score is chosen by the central server, and its gradient is used to update the global model for the current training round. Multi-Krum \cite{krum} is a variation of Krum that balances averaging and Krum. It chooses top-$k$ clients with highest Krum scores. The average of chosen clients' updates is used to update the global model. The prerequisite for Krum to be effective is that the number of malicious clients needs to satisfy $f > (n - 2) / 2$. 

Although the convergence of Krum has been proven in \cite{krum}, authors of Bulyan~\cite{buylan} point out that the attacker can simply deceive Krum to pick the malicious client that converges to an ineffective local minima. Such an attack is launched by manipulating the gradient norms as discussed above. Bulyan refines norm-based aggregation rules such as Krum by adding an extra stage after a client has been chosen by the central server. The added stage is akin to \ac{MeaMed} \cite{generalized-bSGD}. Bulyan first iteratively move clients chosen by Krum or other rules to a candidate set. Once the number of candidates passes the threshold $2f + 3$, Bulyan computes the \ac{MeaMed} on each dimension of candidate gradients. The resulting vector is regarded as the output of Bulyan and subsequently used to update to global model. For Bulyan to be effective, the number of malicious clients needs to satisfy $f > (n - 3) / 4$.

Different from the above approaches, ELITE \cite{elite} uses information gain to filter out malicious updates. ELITE first computes the empirical probability density function for each dimension of gradients, which allows for deriving the dimension-wise information entropy. The sum of all entropy is computed as the total entropy of updates for the current training round. Then for each participating client, their information gain is defined as the difference between the original total entropy and the total entropy with this client being removed. Clients with largest information gains are considered as malicious and hence excluded from the aggregation. The intuition behind ELITE is that benign gradients tend to roughly point at the same direction, namely the direction of the optimal gradient, whereas malicious gradients tend to point at rather different directions. When the majority of clients are benign, removing malicious gradients results in less total entropy as the uncertainty of gradients is reduced.

\subsubsection{Defense Against Free-Rider Attacks}

Since the objective of free-rider attacks is to obtain the global model in the \ac{FL} system, free-rider clients need to upload their own local model such that they can pretend to be benign clients. Free-rider models are constructed with minimum cost. The free-rider can simply upload their received global model to the server \cite{free-rider}, or Gaussian noise may be added to the received model before uploading \cite{gaussian-rider}. The key of defending against free-rider attacks is to identify which clients submit free-rider models. Existing defenses can be categorized into watermarking methods and anomaly detection methods. Watermarking methods incorporate watermark learning tasks on clients, while anomaly detection approaches are learned on the server. If a client model fails to trigger watermarked behaviors or being classified as an anomaly, such client is considered as a free-rider. 

Watermarking neural networks has been studied in the centralized setting \cite{watermarking-1, watermarking-2} to verify the ownership of deep neural networks. Watermarks are commonly embedded into intermediate features or backdoored test samples. In the \ac{FL} scenario, WAFFLE \cite{waffle} is an early work of \ac{FL} watermarking in which the server embeds watermarks by retraining the aggregated model with backdoored samples. However, watermarking on the server side is not suitable for defending against free-rider attacks, as the free-rider model is identical to the global model. FedIPR \cite{fed-ipr} addresses the problem by generating secret watermarks on clients. At the initialization stage of \ac{FL}, FedIPR requires each client to generate their own trigger dataset, watermark embedding matrix and the location of watermarks. In addition to the primary learning task, local models now learns to embed watermarks in both the intermediate features and local trigger set. In the verification stage, client models are fed with their respective trigger set. If the detection error of trigger samples is smaller than a given threshold, this client passes the verification. FedIPR also verifies feature-based watermarks by evaluating the Hamming distance between the watermark in the global model and local secret watermark. One major challenge of FedIPR is that clients may generate conflicting watermarks. Authors of FedIPR proves that different client watermarks can be embedded without conflicts when the total bit-length of watermarks is bounded by the channel number of the global model. If the bit-length exceeds the threshold, FedIPR also gives a lower bound for detecting watermarks.

Anomaly detection based free-rider defense are inspired by anomaly detection approaches in the centralized setting, such as \cite{anomaly-ae, dagmm}. Authors of \cite{gaussian-rider} concatenate client updates on the server to train an auto-encoder. The auto-encoder learns to reconstruct received client updates. In the verification stage, if the reconstruction error induced by updates from one client is larger than then given threshold, this client is deemed as a free-rider. Another approach proposed in \cite{gaussian-rider} is using DAGMM \cite{dagmm} instead of the vanilla auto-encoder. DAGMM detects anomaly data by feeding the latent representation of the auto-encoder to a Gaussian mixture network to estimate the likelihood of the representation being abnormal. 

\section{Model to Data Attacks}

In this section, we will introduce the Model to Data (M2D) attacks in \ac{FL}, which is to reveal a specific attribute, partial or full of the data. We summarized the methods to be non-gradient-based leakage and gradient-based data leakage.

\begin{figure*}[ht!]
\centering
\includegraphics[width=0.7\linewidth]{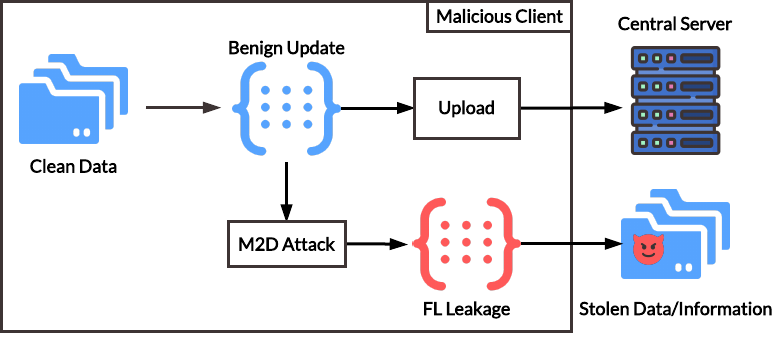}
\caption{\ac{M2D} Attack.}
\label{fig: leakage}
\end{figure*}

\subsection{Non-Gradient-Based Data Leakage}

We define non-gradient-based data leakage as the disclosure of private information that occurs independently of the gradient generated during the training stage. For instance, the leakage can involve identifying specific attributes or membership details within the training data, or recovering original training images from obscured or masked versions. Typically, such leakage exploits the capabilities of a well-trained model to execute these attacks.

\subsubsection{Attribute Inference} 

The paper~\cite{ateniese2015hacking} is one of the earliest works that targets the leakage of private information from an \ac{ML} model. In this paper, the authors construct a novel meta-classifier that is used to attack other \ac{ML} classifiers with the aim of revealing sensitive information from the training data. This is considered a white-box attack, as the adversary has knowledge of both the structure and the parameters of the target model. Specifically, the method assumes full access to a well-trained target model and pre-sets a particular attribute to be identified, determining whether or not it exists in the training data. To do this, the authors first create multiple synthetic training datasets, some of which partially contain the pre-set attributes, while the rest do not. They then train several classification models on these synthetic datasets; the architecture of these classification models is identical to that of the target model. Subsequently, the parameters of these classification models are used as input for training the meta-classifier. Finally, the parameters from the well-trained target model are fed into this meta-classifier to determine if the particular attribute exists in the training data. Both the target model and the meta-classifier are \ac{ML} models, \emph{e.g.}, \ac{ANN}, \ac{HMM}~\cite{baum1966statistical}, \ac{SVM}~\cite{boser1992training}, or \ac{DT}. The authors provide two example cases to evaluate their method. In one example, they identify the speaker's nationality using a speech recognition dataset processed by an \ac{HMM}. Later, they use an \ac{SVM} to set up a network traffic classifier to distinguish between two kinds of traffic conditions, using the meta-classifier to identify the type of traffic. In both examples, the meta-classifiers are \ac{DT}s.

\subsubsection{Membership Identification} 

The above work is further improved by~\cite{shokri2017membership}, who focus on membership identification attacks. They propose a shadow training technique to identify whether specific samples are part of the training dataset. The membership inference problem is formulated as a classification task. An attack model is trained to distinguish between the behavior of shadow models when fed with forged training data. These shadow models are designed to behave similarly to the target model. The approach qualifies as a black-box attack, meaning that the attacker only possesses knowledge of the output for a given input. Several effective methods have been developed for generating forged training data for the shadow models. The first method utilizes black-box access to the target model to synthesize the data. The second method leverages statistical information related to the target model's training dataset. In the third method, it is assumed that the adversary has access to a noisy version of the target model's training dataset. While the first method operates without assuming any prior knowledge about the distribution of the target model's training data, the second and third methods allow the attacker to query the target model just once before determining whether a particular record was part of its training dataset.

\subsubsection{Image Recovery} 

In terms of recovering valuable information from obfuscated images, \cite{mcpherson2016defeating} is one of the earliest works to the best of our knowledge. Obfuscated images are easily accessible through various data protection techniques (\emph{e.g.}, blur, mask, corrupt, and P3)~\cite{carrell2013hiding,li2019hideme}. In the study~\cite{mcpherson2016defeating}, the authors utilized a \ac{DL} model to recover valuable information from obfuscated images for classification tasks. They assumed that the adversary has access to a portion of the original training data and applied one of the encryption methods to those images to train the attack model. For this reason, their method is generally not suitable for most real-world scenarios.

To demonstrate how neural networks can overcome privacy protection measures, they employed four commonly used datasets for recognizing faces, objects, and handwritten digits. Each of these tasks carries substantial privacy concerns. For instance, the successful identification of a face could infringe upon the privacy of an individual featured in a captured video. Recognizing digits could enable the deduction of written text content or vehicular registration numbers.

The final results are impressive. On the MNIST~\cite{lecun1998mnist} dataset, they achieved an accuracy of about 80\% for images encrypted by P3 with a recommended threshold level of 20. Conversely, the accuracy exceeds 80\% when the images are masked by windows of resolution $8 \times 8$. On the CIFAR-10~\cite{krizhevsky2009learning} dataset, only vehicle and animal images were used for experiments, achieving an accuracy of 75\% against P3 with a threshold of 20. When deploying a $4 \times 4$ mask on the images, the accuracy is approximately 70\%, and it drops to 50\% when masking with $8 \times 8$ resolution. On the AT\&T~\cite{at&t1994database} dataset, the proposed method achieved a remarkable accuracy of 97\% against P3 with a threshold of 20, over 95\% against various mask sizes, and 57\% against face blurring. On the FaceScrub~\cite{ng2014datadriven} dataset, they achieved an accuracy of 57\% against masking the face with a $16 \times 16$ window and 40\% against P3 with a threshold of 20.

\begin{figure}[ht!]
    \centering
    \includegraphics[width=0.7\linewidth]{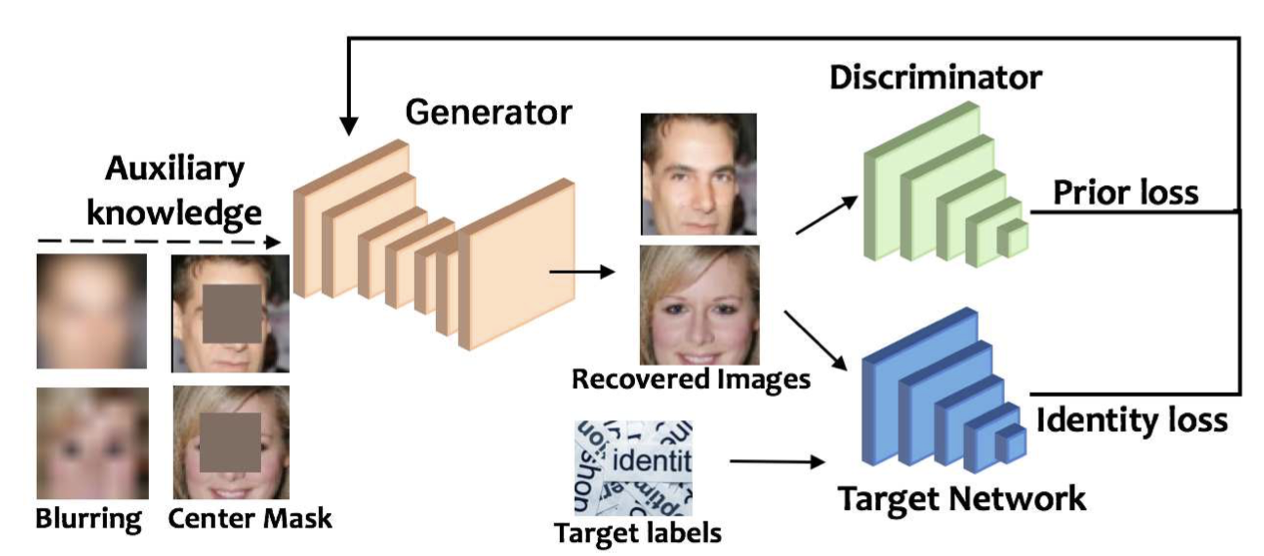}
    \caption{Overview of the \ac{GMI} attack method. \cite{zhang2020secret}}
    \label{fig:gmi}
\end{figure}

In more recent work~\cite{zhang2020secret}, the authors utilize a \ac{GAN}, trained on a public dataset, to recover missing sensitive regions in images; this is termed the \ac{GMI} attack, as shown in Figure~\ref{fig:gmi}. A diversity loss is proposed to encourage diversity in the images synthesized by the generator when projected into the target network's feature space. This is essential during the training of the \ac{GAN} on the public dataset because the adversary aims for the generated images to be distinct in the feature space of the target model. If different images map to the same feature space, the adversary cannot discern which generated image corresponds to the private data's features, thus failing to reveal the private information.

The authors assume that the adversary has access to the well-trained target model, which serves as a discriminator, as well as to the target label of the input corrupted image. Initially, the generator is used to create an image, which is then fed into two separate discriminators to calculate the prior loss and identity loss. In subsequent rounds, these two losses, along with the corrupted image, are used as inputs for the generator to produce the next iteration of the reconstructed image. Upon completing the training of the \ac{GAN}, the adversary, during the reveal phase, only needs to continue optimizing the generator's inputs so that the generated images are sufficiently realistic while also maximizing likelihood in the target model.

The datasets employed for evaluation are MNIST~\cite{lecun1998mnist}, ChestX-ray8~\cite{wang2017chestx}, and CelebA~\cite{liu2015deeplearningface}. The experimental results indicate that without using the corrupted image as an input for the generator, the attack's success rate is approximately 28\%, 44\%, and 46\% on target networks VGG-16~\cite{simonyan2014verydeepconvolutional}, ResNet-152~\cite{he2016deepresiduallearning}, and face.evoLVe~\cite{cheng2017knowyouone}, respectively. However, when the corrupted image is incorporated, the accuracy increases to 43\%, 50\%, and 51\% for blurred input images; 78\%, 80\%, and 82\% for center-masked images; and 58\%, 63\%, and 64\% for face T-masked images. Consequently, the inclusion of corrupted images as auxiliary information has a significant impact on the attack's accuracy.

\subsection{Gradient-Based Data Leakage}

Concerning gradient-based data leakage, this refers to techniques that exploit gradients from the target model to expose privacy-sensitive information. \ac{DL} models are trained on datasets, and parameter updates occur through alignment with the feature space. This establishes an inherent relationship between the weights or gradients and the dataset. Consequently, numerous studies aim to reveal private information by leveraging these gradients. The effectiveness and success rates of gradient-based approaches have consistently surpassed those of non-gradient-based methods. Unlike non-gradient-based leakage, gradient-based data leakage can occur even in models that have not yet converged.

\subsubsection{Partial Recovery} 

Hitaj \emph{et al.}~\cite{hitaj2017deep} proposed a data recovery method that utilizes a trained victim model and a target label. The method aims to generate new data closely resembling the distribution of the training dataset. This attack is formulated as a generative process using a \ac{GAN}. In a \ac{FL} system, an attacker can pose as a participant to reveal private data from the victim by modeling the feature space. Suppose the attacker masquerades as a malicious participant with a portion of training samples that have correct labels, along with a portion of samples generated via \ac{GAN} with incorrect labels. The attacker's goal is to produce a dataset that shares the same feature distribution as the other participants, leveraging \ac{GAN} and the global gradients downloaded from the parameter server. 

In Algorithm~\ref{alg:hitaj2017deep}, the victim trains its local model on its own dataset for several iterations until it achieves an accuracy beyond a preset threshold. Subsequently, the malicious actor uses the updated local model as the discriminator. The weights in the discriminator are fixed, and a generator is trained to maximize the confidence of a specific class. This is an indirect data recovery method, sensitive to the variance in the victim's training data~\cite{melis2019exploiting}. Although the generated images are consistent with the data distribution, they do not correspond to the actual training dataset. In other words, the generated images cannot be mapped back to the training data.

Another related work by \ac{GGL}~\cite{li2022auditing} also employs a \ac{GAN} to generate fake data. In this approach, the weights of the \ac{GAN} are pretrained and fixed, while the trainable parameters in \ac{GGL} are the input sequences to the \ac{GAN}. The label inference part is adapted from \ac{iDLG}~\cite{zhao2020idlg}, requiring a batch size of 1. Unlike other methods, \ac{GGL} uses \ac{CMA-ES} and \ac{BO} as optimizers to reduce the variability in the generated data. Although the data generated by \ac{GGL} is not identical to true data, it is sufficiently similar (see Table~\ref{tab:ggl-results}), providing \ac{GGL} with robustness against various defense strategies like gradient noising, clipping, or compression. The generated images are influenced by two factors: 1) the inferred ground-truth label, which specifies the image classification, and 2) fine-tuning based on gradient information to make the image as similar as possible to the true image.

\begin{table*}[!ht]
\begin{center}
\caption{Typical experimental results performed on \ac{GGL} are shown below. The backbone network is \emph{ResNet-18} and the dataset is ILSVRC2012 with a resolution of $256*256$. \cite{ren2023gradient}}
\label{tab:ggl-results}
\begin{tabular}{c c | c c c c}

&\textbf{Ground True} & \multicolumn{4}{c}{Generated Images} \\

\multicolumn{1}{m{0.12cm}}{\rotatebox{90}{\textbf{black grouse}}} & \makecell*[c]{\includegraphics[width=0.12\linewidth]{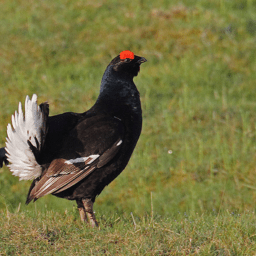}} & \makecell*[c]{\includegraphics[width=0.12\linewidth]{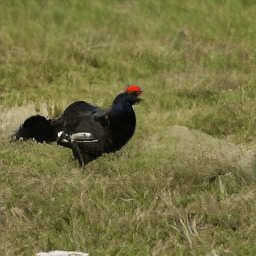}} & \makecell*[c]{\includegraphics[width=0.12\linewidth]{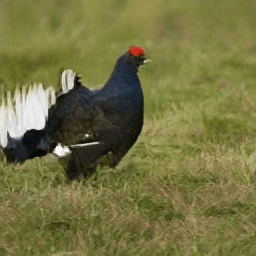}} & \makecell*[c]{\includegraphics[width=0.12\linewidth]{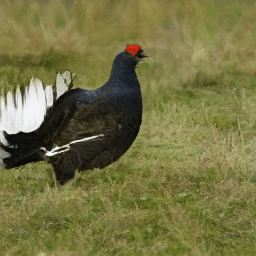}} & \makecell*[c]{\includegraphics[width=0.12\linewidth]{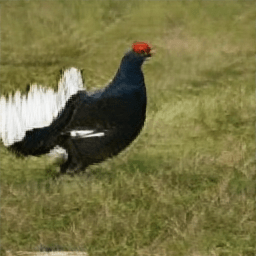}} \\

\multicolumn{1}{m{0.1cm}}{\rotatebox{90}{\textbf{tiger beetle}}} & \makecell*[c]{\includegraphics[width=0.12\linewidth]{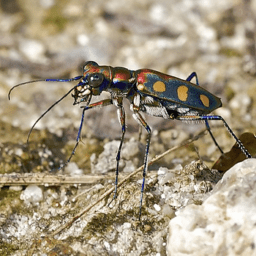}} & \makecell*[c]{\includegraphics[width=0.12\linewidth]{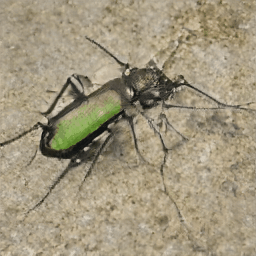}} & \makecell*[c]{\includegraphics[width=0.12\linewidth]{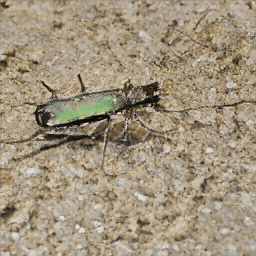}} & \makecell*[c]{\includegraphics[width=0.12\linewidth]{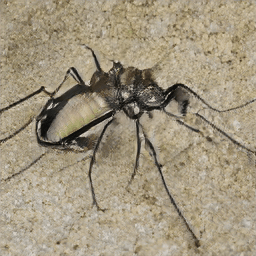}} & \makecell*[c]{\includegraphics[width=0.12\linewidth]{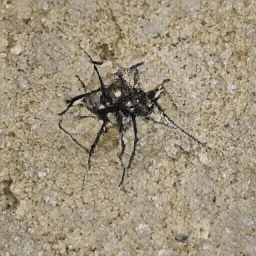}} \\

\multicolumn{1}{m{0.1cm}}{\rotatebox{90}{\textbf{cliff dwelling}}} & \makecell*[c]{\includegraphics[width=0.12\linewidth]{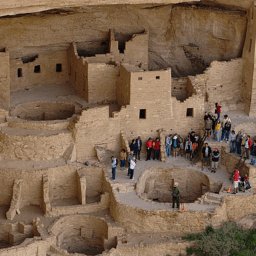}} & \makecell*[c]{\includegraphics[width=0.12\linewidth]{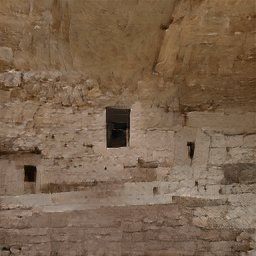}} & \makecell*[c]{\includegraphics[width=0.12\linewidth]{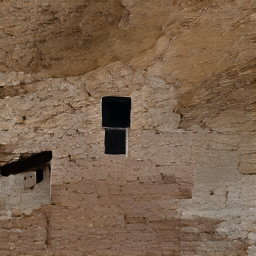}} & \makecell*[c]{\includegraphics[width=0.12\linewidth]{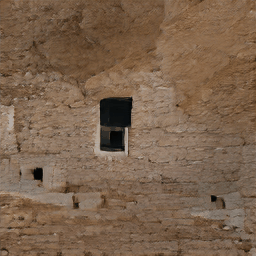}} & \makecell*[c]{\includegraphics[width=0.12\linewidth]{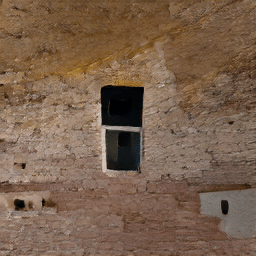}} \\

\multicolumn{1}{m{0.1cm}}{\rotatebox{90}{\textbf{basset hound}}} & \makecell*[c]{\includegraphics[width=0.12\linewidth]{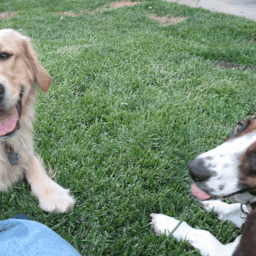}} & \makecell*[c]{\includegraphics[width=0.12\linewidth]{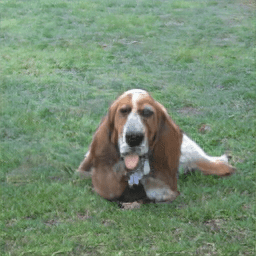}} & \makecell*[c]{\includegraphics[width=0.12\linewidth]{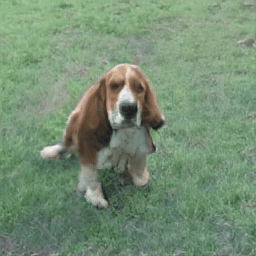}} & \makecell*[c]{\includegraphics[width=0.12\linewidth]{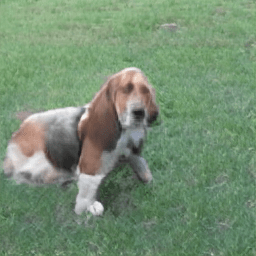}} & \makecell*[c]{\includegraphics[width=0.12\linewidth]{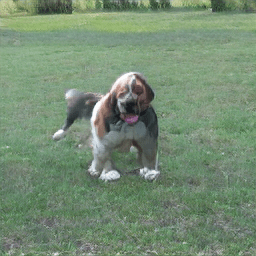}} \\

\multicolumn{1}{m{0.1cm}}{\rotatebox{90}{\textbf{sweatshirt}}} & \makecell*[c]{\includegraphics[width=0.12\linewidth]{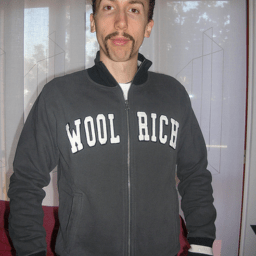}} & \makecell*[c]{\includegraphics[width=0.12\linewidth]{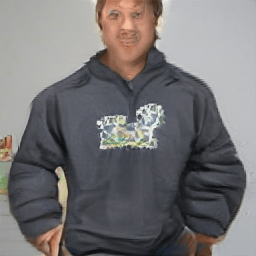}} & \makecell*[c]{\includegraphics[width=0.12\linewidth]{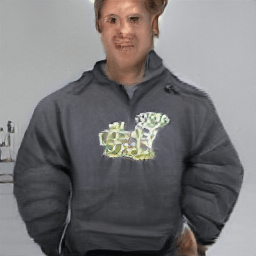}} & \makecell*[c]{\includegraphics[width=0.12\linewidth]{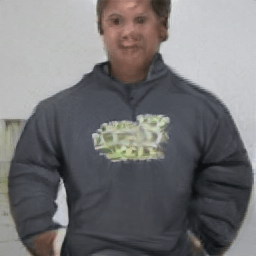}} & \makecell*[c]{\includegraphics[width=0.12\linewidth]{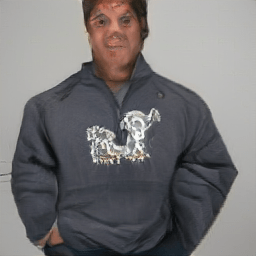}} \\

\multicolumn{1}{m{0.1cm}}{\rotatebox{90}{\textbf{radiator grille}}} & \makecell*[c]{\includegraphics[width=0.12\linewidth]{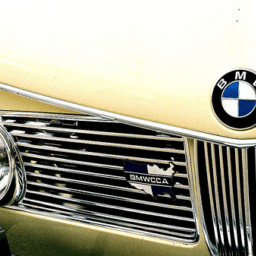}} & \makecell*[c]{\includegraphics[width=0.12\linewidth]{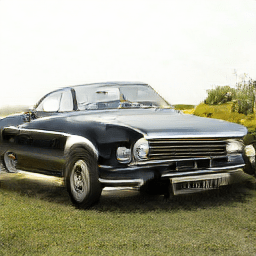}} & \makecell*[c]{\includegraphics[width=0.12\linewidth]{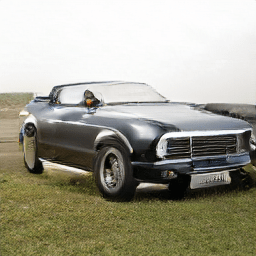}} & \makecell*[c]{\includegraphics[width=0.12\linewidth]{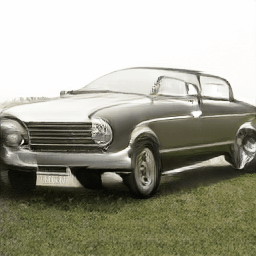}} & \makecell*[c]{\includegraphics[width=0.12\linewidth]{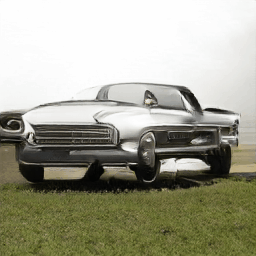}} \\

\multicolumn{1}{m{0.1cm}}{\rotatebox{90}{\textbf{pig}}} & \makecell*[c]{\includegraphics[width=0.12\linewidth]{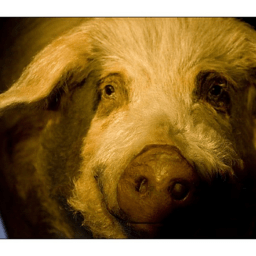}} & \makecell*[c]{\includegraphics[width=0.12\linewidth]{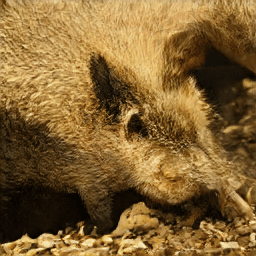}} & \makecell*[c]{\includegraphics[width=0.12\linewidth]{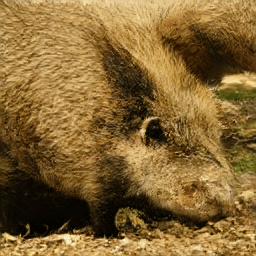}} & \makecell*[c]{\includegraphics[width=0.12\linewidth]{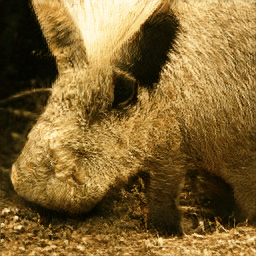}} & \makecell*[c]{\includegraphics[width=0.12\linewidth]{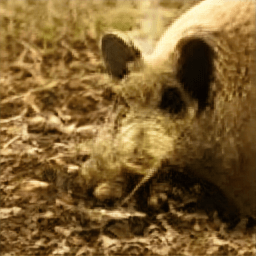}} \\
\end{tabular}
\end{center}
\end{table*}

\begin{algorithm}[ht!]
\renewcommand{\algorithmicensure}{\textbf{Assume:}}
\caption{The proposed work from \cite{hitaj2017deep}}
    \begin{algorithmic}[1]
    \ENSURE two participants V and M who have common learning goals.
    \REQUIRE V's local dataset $D_v$ with label $L_a$ and $L_b$. \\\qquad\quad
     M's local dataset $D_m$ with label $L_b$ and $L_c$.
     \end{algorithmic}
    \textbf{a. Parameter Server}
    \begin{algorithmic}[1]
    \STATE build model and initialize weights.
    \STATE send the initial weights to the clients.
    \STATE local training on victim and malicious clients.
    \STATE receive the trained local weights and generate the global model.
    \STATE repeat Step 2 and 3 until the model converges.
    \end{algorithmic}

    \textbf{b. Victim Client}
    \begin{algorithmic}[1]
    \STATE download the global weights from parameter server.
    \STATE train the local model on its local dataset $D_v$.
    \STATE upload the local model to the parameter server.
    \end{algorithmic}

    \textbf{c. Malicious Client}
    \begin{algorithmic}[1]
    \STATE download the global weights from parameter server.
    \STATE train a GAN model to generate fake data of class $L_a$.
    \STATE generate many fake data using GAN and relabel them with $L_c$ to update the local dataset $D_m$.
    \STATE train the local model on the updated local dataset $D_m$.
    \STATE upload the local model to the parameter server.
    \end{algorithmic}
\label{alg:hitaj2017deep}
\end{algorithm}

\subsubsection{Full Recovery (Discriminative)} 

\begin{figure*}[ht!]
    \centering
    \includegraphics[width=0.99\linewidth]{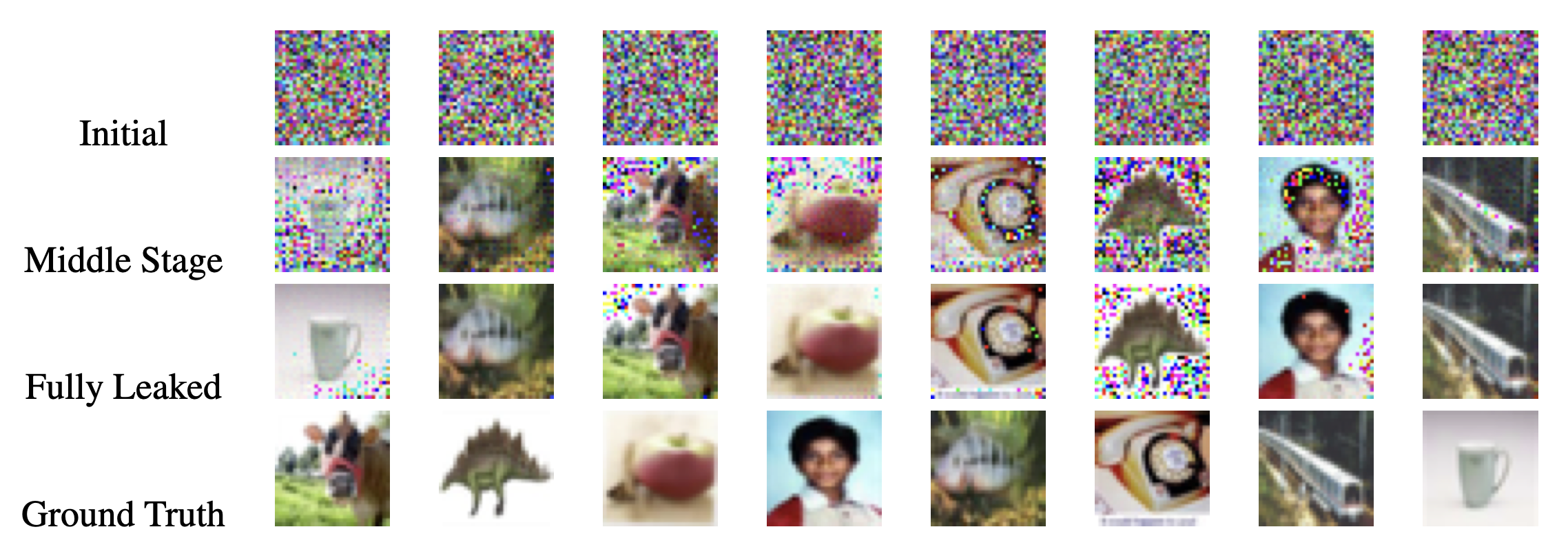}
    \caption{Although the sequence might differ and additional artifact pixels are present, deep leakage in batched data still generates images that closely resemble the original versions. \cite{zhao2020idlg}}
    \label{fig:dlg-results}
\end{figure*}

Zhu \emph{et al.}~\cite{zhu2019deep} introduced \ac{DLG}, framing the image recovery task as a regression problem. Initially, the shared local gradient is derived from a victim participant, and a batch of ``dummy'' images and labels is randomly initialized. These are then used to calculate the ``dummy'' gradient through standard forward-backward propagation, employing the L-BFGS optimizer~\cite{liu1989limited}. This process leverages regression techniques to decipher intricate patterns within the gradient, thereby reconstructing the private image data. The approach provides a powerful framework for \ac{M2D} attacks. Importantly, it is the input ``dummy'' data that is updated---not the model parameters---by minimizing the \ac{MSE} between the ``dummy'' gradient and the shared local gradient. This strategy prioritizes the fidelity of the reconstructed image, ensuring preservation of essential features and details. Among existing leakage methods, \ac{DLG} is unique in achieving precise pixel-wise data revelation without requiring additional information. The technique is innovative and deploys unique algorithms to achieve an unparalleled level of precision. Some results from \ac{DLG} of batch data are provided in Figure~\ref{fig:dlg-results}. It marks a significant advancement in the field of gradient leakage, opening new avenues for research and application. Although \ac{DLG} can perform attacks on multiple images simultaneously, the accuracy in label inference remains suboptimal. This limitation is an active area of research, with ongoing efforts to improve label inference accuracy without compromising image recovery fidelity. In conclusion, \ac{DLG} offers a novel approach to image recovery, utilizing groundbreaking algorithms to attain high precision. Its potential applications extend far beyond existing methods, positioning it at the forefront of technological advancements in the field.

Zhao \emph{et al.}~\cite{zhao2020idlg} introduced a novel method known as \ac{iDLG}, which focuses on the identification of labels in a more accurate manner. This technique involves calculating the derivative of the cross-entropy loss with respect to one-hot labels for each class in the classification task. The crux of this approach lies in the distinct ranges of the derivative values that correspond to different labels. The authors discovered that the derivative value for the ground-truth label uniquely falls within the range of [-1, 0], while the derivatives corresponding to incorrect labels lie within the range of [0, 1]. This separation of value ranges provides a solid basis for identifying the correct label. By simply examining the derivative value, the system can distinguish the correct label from incorrect ones. However, this method has a limitation concerning the batch size: the batch size must not exceed 1 during the process. While this constraint may affect efficiency in large-scale applications, the \ac{iDLG} method's unique approach to label identification through derivative analysis represents a significant contribution to the field of gradient leakage. It opens avenues for future research to potentially refine this technique and mitigate its limitations.

\begin{figure}[ht!]
    \centering
    \includegraphics[width=0.8\linewidth]{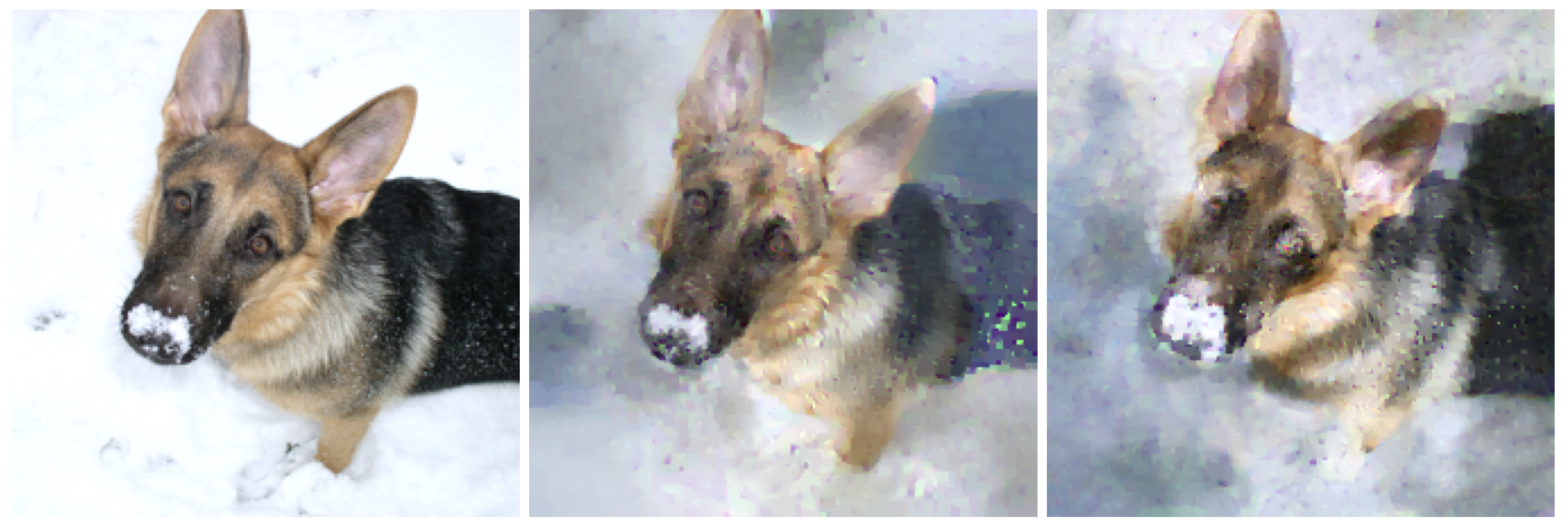}
    \caption{Reconstructed image using its gradient features. On the left is the ground true image taken from the validation dataset. The center image is reconstructed using a trained ResNet-18 model that has been trained on ILSVRC2012 dataset. On the right is the image rebuilt using a trained ResNet-152 model. \cite{geiping2020inverting}}
    \label{fig:ig-results}
\end{figure}

In addition to the low accuracy of label inference, \ac{DLG} often fails to recover the image from the gradient when the data variance is large, see Figure~\ref{fig:ig-results}. This is particularly common for datasets with a large number of classes. \ac{IG}~\cite{geiping2020inverting} improved the stability of \ac{DLG} and \ac{iDLG} by introducing a magnitude-invariant cosine similarity metric for the loss function, termed \ac{CD}. This approach aims to find images that yield similar prediction changes in the classification model, rather than images that produce closely matching values with a shared gradient. The method demonstrates promising results in recovering high-resolution images (\emph{i.e.}, \(224 \times 224\)) when trained with large batch sizes (\emph{i.e.}, \(\#Batch = 100\)); however, the \ac{PSNR} remains unacceptably low.

Similar to \cite{geiping2020inverting}, Jeon \emph{et al.}~\cite{jeon2021gradient} argued that relying solely on gradient information is insufficient for revealing private training data. They introduced GIAS, which employs a pre-trained model for data revelation. Yin \emph{et al.}~\cite{yin2021see} reported that in image classification tasks, the ground-truth label can be easily inferred from the gradient of the last fully-connected layer. Additionally, \ac{BN} statistics can significantly improve the efficacy of gradient leakage attacks and facilitate the revelation of high-resolution private training images.

Another approach to gradient leakage attacks is based on generative models. Wang \emph{et al.}~\cite{wang2019beyond} trained a \ac{GAN} with a multi-task discriminator, named mGAN-AI, to generate private information based on gradients.

\subsubsection{Full Recovery (Generative)} 

\begin{figure}[th!]
    \centering
    \includegraphics[width=0.65\linewidth]{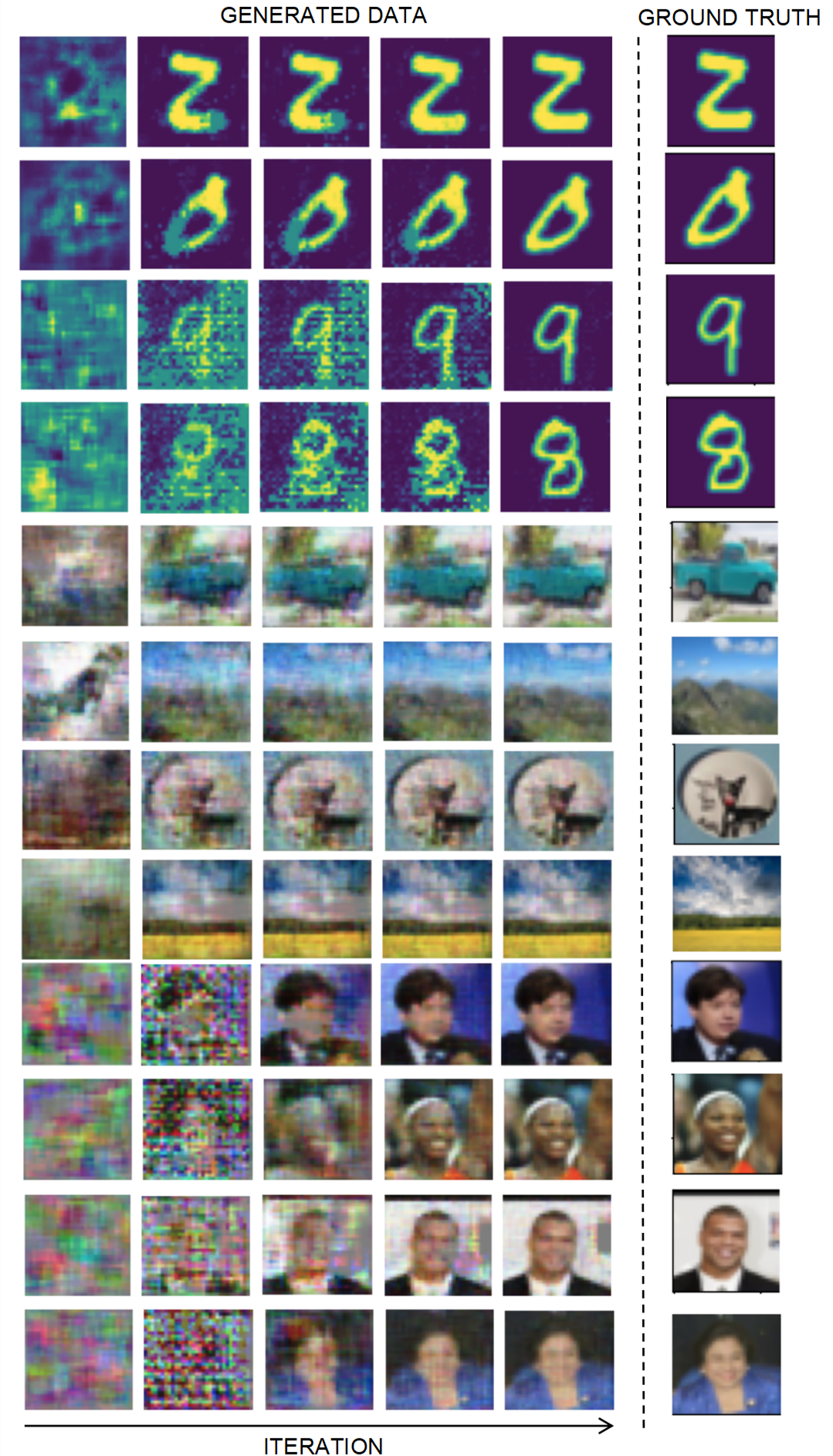}
    \caption{Examples of data leakage attack using the \ac{GRNN} on the global model. \cite{ren2022grnn}}
    \label{fig:grnn-results}
\end{figure}

In the work~\cite{ren2022grnn}, the \ac{GRNN} was proposed as a method for reconstructing private training data along with its associated labels. The model is capable of handling large batch sizes and high-resolution images. Some examples are provided in Figure~\ref{fig:grnn-results} Inspired by both \ac{GAN} and \ac{DLG} methods, \ac{GRNN} introduces a gradient-driven approach for image creation that effectively addresses the challenges of stability and data quality commonly associated with \ac{DLG} methodologies.

The novel \ac{GRNN}, which serves as an innovative data leakage attack technique, is capable of retrieving private training images with resolutions up to $256 \times 256$ and batch sizes of $256$. This makes it particularly well-suited for \ac{FL} applications, as both the local gradient \(g\) and the global model \(\mathcal{F}(\bullet)\) are easily accessible within the system's configuration. The \ac{GRNN} algorithm employs a dual-branch structure to generate fake training data \(\hat{x}\) and corresponding labels \(\hat{y}\). It is trained to estimate a fake gradient \(\hat{g}\), computed from the generated data \(\hat{x}\) and labels \(\hat{y}\), such that it closely matches the true gradient \(g\) associated with the global model. The divergence \(\mathcal{D}\) between the true and fake gradients is evaluated using a combination of \ac{MSE}, \ac{WD}, and \ac{TVLoss} metrics.

Through empirical testing on various image classification challenges, the \ac{GRNN} approach has been rigorously compared to cutting-edge alternatives, showing significantly better results across multiple metrics. The trial findings confirm that the proposed method is notably more stable and capable of generating images of superior quality, especially when applied to large batch sizes and high resolutions.

Compared to the most latest work~\cite{zhu2019deep,zhao2020idlg,geiping2020inverting}, \ac{GRNN} takes a generative approach, which shows high stability for recovering high-resolution images (\emph{i.e.} up to $256 \times 256$) with a large batch size (\emph{i.e.} $\#Batch=256$). Table~\ref{tab:compareMethod} presents the key differences between \ac{DLG}, \ac{iDLG} \ac{IG} and \ac{GRNN}.

\begin{algorithm}[ht!]
    \caption{GRNN: Data Leakage Attack~\cite{ren2022grnn}}
    \begin{algorithmic}[1]
    \STATE $g \leftarrow \partial \mathcal{L}(\mathcal{F}(<x, y>, \theta)) / \partial \theta$; \hfill \#Produce true gradient on local client.
    \STATE $v \leftarrow$ Sampling from $\mathcal{N}(0, 1)$; \hfill \#Initialize random vector inputs.
    \FOR{each iteration $i \in [1,2,...,I]$}
        \STATE $(\hat{x}_i, \hat{y}_i) \leftarrow \mathcal{G}(v | \hat{\theta}_i)$; \hfill \#Generate fake images and labels.
        \STATE $\hat{g}_i \leftarrow {\partial \mathcal{L}(\mathcal{F}(<\hat{x}_i, \hat{y}_i>, \theta))}/{\partial \theta}$; \hfill \#Get fake gradient on global model.
        \STATE $\mathcal{D}_i \leftarrow \hat{\mathcal{L}}(g, \hat{g}_i, \hat{x}_i)$; \hfill \#Loss between true and fake gradient.
        \STATE $\hat{\theta}_{i+1} \leftarrow {\hat{\theta}_i - \eta (\partial \mathcal{D}_i} / {\partial \hat{\theta}_i)}$; \hfill \#Update \ac{GRNN} model.
    \ENDFOR
    \STATE \textbf{return} $(\hat{x}_I, \hat{y}_I)$; \hfill \#Return generated fake images and labels.
    \end{algorithmic}
    \label{alg:InferenceAttack}
\end{algorithm}

\begin{table*}[ht!]
    \setlength{\tabcolsep}{10pt}
    \renewcommand{\arraystretch}{1.5}
    \centering
    \caption{Comparison of different related works on gradient leakage.~\cite{ren2022grnn}}
    
    \begin{tabular}{c|c|c|c|c}
         \hline
         \textbf{Method} & \textbf{\makecell[c]{Recovery\\ Mode}} & \textbf{\#Batch} & \textbf{Resolution} & \textbf{\makecell[c]{Loss Function}} \\
         \hline
         \ac{DLG}\cite{zhu2019deep} & Discriminative & Small, up to 8 & Low $64\times64$ & \ac{MSE}  \\
         \hline
         \ac{iDLG}\cite{zhao2020idlg} & Discriminative & Small, only 1 & Low $64\times64$ & \ac{MSE}  \\
         \hline
         \ac{IG}\cite{geiping2020inverting} & Discriminative & Medium, up to 100 & High $224\times224$ & \makecell[c]{\ac{CD} \& \ac{TVLoss}}  \\
         \hline
         \ac{GGL}\cite{li2022auditing} & Generative & Small, only 1 & High $224\times224$ & \makecell[c]{\ac{CMA-ES} \& \ac{BO}} \\
         \hline
         \ac{GRNN}\cite{ren2022grnn} & Generative & Large, up to 256 & High $256\times256$ & \makecell[c]{\ac{MSE} \& \ac{WD} \& \ac{TVLoss}} \\
         \hline
    \end{tabular}
    \label{tab:compareMethod}
\end{table*}
\subsection{Defense Against M2D Attacks}


The issue of \ac{M2D} attack methods has garnered significant attention in the world of \ac{ML} and \ac{DL}. This issue has sparked concern as it can lead to the unintended exposure of information. In response, numerous methods and techniques have been proposed to understand, mitigate, and control this leakage, \emph{e.g.}, gradient perturbation~\cite{zhu2019deep, yang2020accuracy, sun2020ldp, sun2021soteria, ren2022grnn}, data obfuscation or sanitization~\cite{hasan2016effective, chamikara2018efficient, chamikara2020efficient, lee2021digestive, chamikara2021privacy}, and other methods~\cite{bu2020deep, ren2020fedboost, li2020privacy, yadav2020differential, wei2021gradient, ren2023gradient}. These methods aim to limit the extent of information that can be exposed, ensuring that models operate with the requisite confidentiality and integrity. Defense against \ac{M2D} attacks has emerged as a compelling and dynamic research area within the field. \ac{M2D} attacks involve malicious attempts to extract or manipulate sensitive information directly from the data used in training models. This field of research explores various strategies and mechanisms to shield against these attacks, preserving the privacy of the data and maintaining the robustness of the models.

Numerous measures have been undertaken to safeguard personal data against the \ac{M2D} attack. Techniques such as gradient perturbation, data obfuscation or sanitization, \ac{DP}, \ac{HE}, and \ac{MPC} are among the most prominent methods for ensuring the privacy of both the private training data and the publicly shared gradient exchanged between the client and server. Experiments conducted by Zhu \emph{et al.}~\cite{zhu2019deep} focused on two specific noise types: Gaussian and Laplacian. Their findings revealed that the key factor affecting the outcome was the magnitude of the distribution variance, rather than the type of noise itself. When the variance exceeds \(10^{-2}\), the leakage attack fails; concurrently, there is a significant decline in the model's performance at this variance level. Chamikara \emph{et al.}~\cite{chamikara2021privacy} introduced a technique for perturbing data, affirming that this approach maintains model performance without compromising the confidentiality of the training data. In this context, the dataset is treated as a data matrix, and a multidimensional transformation is applied to project it into a new feature space. Various degrees of transformation are used to perturb the input data, guaranteeing an adequate level of alteration. A central server is responsible for creating global perturbation parameters in this technique. Notably, a potential drawback is that the perturbation process could distort the architectural structure of image-related data. Wei \emph{et al.}~\cite{wei2021gradient} employed \ac{DP} to introduce noise into the training datasets of each client and formulated a per-example-based \ac{DP} method known as Fed-CDP. They developed a dynamic decay noise injection strategy to improve both inference performance and the level of gradient leakage defense. Nevertheless, experimental findings indicate that, despite successfully hindering the reconstruction of training data from the gradient, this method leads to a considerable decline in inference accuracy. Additionally, since \ac{DP} is applied to every training instance, the computational overhead becomes substantial.

When computing the gradient, \ac{PRECODE}~\cite{scheliga2022precode} aims to prevent the input information from propagating through the model. \ac{PRECODE} introduces a module before the output layer to transform the latent representation of features using a probabilistic encoder-decoder. This encoder-decoder is comprised of two fully-connected layers. The first layer encodes the input features into a sequence and then normalizes this sequence based on calculated mean and standard deviation values. The mean is computed from the first half of the sequence, while the standard deviation is derived from the remaining half. Finally, the decoder translates the normalized sequence back into a latent representation, which then serves as input to the output layer. This normalization step between the encoder and decoder prevents the input information from affecting the gradient, thereby allowing \ac{PRECODE} to resist the leakage of input information through the gradient. However, the insertion of two fully-connected layers in front of the output layer results in a significant computational cost. This is why only three very shallow neural networks were used for experiments in their paper.

Recent studies have uncovered that shared gradients can result in the potential exposure of sensitive data, leading to privacy violations. The work in \cite{ren2023gradient} presents an exhaustive examination and offers a fresh perspective on the issue of gradient leakage. These theoretical endeavors have culminated in the development of an innovative gradient leakage defense strategy that fortifies any model architecture by implementing a private key-lock mechanism. The only gradient communicated to the parameter server for global model aggregation is the one that has been secured with this lock. The newly formulated learning approach, termed FedKL, is designed to withstand attacks that attempt to exploit gradient leakage.

The key-lock component has been meticulously designed and trained to ensure that without access to the private details of the key-lock system: a) the task of reconstructing private training data from the shared gradient becomes unattainable, and b) there is a considerable deterioration in the global model's ability to make inferences. The underlying theoretical reasons for gradients potentially leaking confidential information are explored, and a theoretical proof confirming the efficacy of our method is provided.

The method's robustness has been verified through extensive empirical testing across a variety of models on numerous widely-used benchmarks, showcasing its effectiveness in both maintaining model performance and protecting against gradient leakage.

In the study~\cite{ren2023gradient}, a theoretical foundation is laid to demonstrate that the feature maps extracted from the fully-connected layer, convolutional layer, and \ac{BN} layer contain confidential details of the input data. These details are not only encompassed within the feature maps but also coexist within the gradient during the process of backward propagation. Furthermore, it is posited that gradient leakage attacks can only succeed if there is adequate alignment between the gradient spaces of the global and local models.

As a solution, they proposed FedKL, a specialized key-lock module that excels at differentiating, misaligning, and safeguarding the gradient spaces using a private key. This is accomplished while preserving federated aggregation comparable to conventional \ac{FL} schemes. Specifically, the operations of scaling and shifting in the normalization layer are restructured. A private key, generated randomly, is fed into two fully-connected layers. The resulting outputs function as exclusive coefficients for the scaling and shifting procedures. Both theoretical analysis and experimental results affirm that the proposed key-lock module is efficient and effective in protecting against gradient leakage attacks. This is achieved by masking the uniformity of confidential data in the gradient, thus making it challenging for a malicious attacker to perform forward-backward propagation in the absence of the private key and the lock layer's gradient. Consequently, the task of approximating the shared gradient in the \ac{FL} framework to reconstruct local training data becomes unachievable.

\section{Composite Attacks}

\begin{table}[ht!]
\renewcommand{\arraystretch}{1.5}
\centering
\caption{Characteristics of Composite Attacks}
\begin{tabular}{ c  c } 
\hline
\textbf{Name of Attack} & \textbf{Distinctive Feature} \\
\hline
\hline
Direct Boosting \cite{adversarial-lens} & boosting malicious updates \\ 
\hline
Separated Boosting \cite{adversarial-lens} & regularized update boosting \\
\hline
Model Replacement \cite{how-to-bd} & replace converging global model \\
\hline
PGD \cite{pgd} & bounded update projection \\
\hline
Edge case + PGD \cite{yes-we-can} & PGD on minority samples \\
\hline
Median Interval \cite{little-stat} & \makecell[c]{median cheating with \\ normalized updates} \\
\hline
DBA\cite{dba} & distributed backdoor trigger \\
\hline
TrojanDBA \cite{trojan-dba} & distributed and learnable trigger \\
\hline
Neurotoxin \cite{neurotoxin} & \makecell[c]{tampering insignificant \\ model weights} \\
\hline
RL Neurotoxin \cite{rl-neurotoxin} & \makecell[c]{searching Neurotoxin \\ parameters with RL} \\
\hline
F3BA \cite{f3ba} & \makecell[c]{sign-flipping on \\ insignificant weights} \\
\hline
\makecell[c]{Rare Word \\ Embedding} \cite{rare-word-embedding} & \makecell[c]{tampering stale \\ word embeddings} \\
\hline
\makecell[c]{Future Update \\ Approximation} \cite{two-heads} & \makecell[c]{estimating future updates \\ from malicious clients} \\
\hline
Sudden Collapse \cite{apa} & \makecell[c]{estimating potent \\ malicious gradients} \\
\hline
\end{tabular}
\label{tab:composite_attack}
\end{table}

We define composite attacks as threat models that corrupt multiple aspects of \ac{FL}. The attacker can combine \ac{D2M} and \ac{M2M} attacks to launch backdoor attacks. The attacker surreptitiously adds trigger patterns to local training data, then poisons model updates such that the global model learns how to react to triggers. Backdoored models behave normally when fed with clean data. In the presence of trigger data, these models are trained to give predictions designated by the attacker. 

\begin{figure*}[h!]
\label{fig:triggers}
\centering
\includegraphics[width=0.8\linewidth]{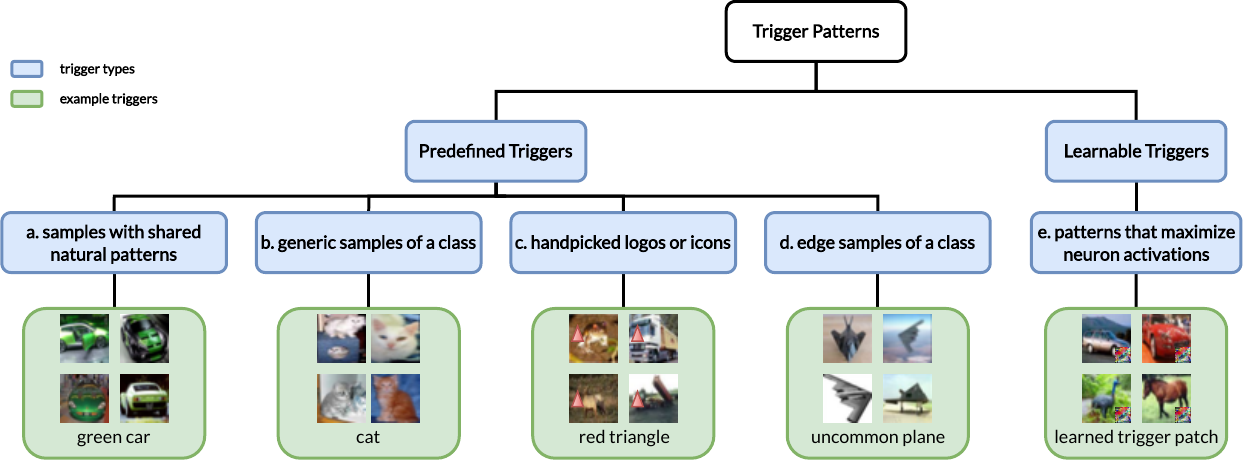}
\caption{An Overview of Trigger Patterns. Among these trigger types, a and b are mostly associated with label-flipping. Type c is a common strategy for injecting triggers into arbitrary samples. Type d uses samples at the tail of the data distribution to induce erroneous predictions for underrepresented data. Type d appears in more recent studies.}
\end{figure*}

Trigger patterns vary from one attack to the other. We summarize existing triggers in Figure~\ref{fig:triggers}. Generic samples of a class or samples with shared patterns are commonly used in label-flipping attacks, these attacks can be further enhanced by incorporating \ac{M2M} attacks. Triggers based on certain natural patterns are also known as semantic triggers \cite{how-to-bd} . Handpicked logos or icons are common trigger patterns for backdoor injection. Edge samples, namely samples at the tail of the data distribution, are used in attacks targeting underrepresented data, which can significantly damage the fairness for the minority group. Lastly, learnable triggers is a relatively new strategy appears in recent studies. 

\begin{figure*}[h!]
\centering
\includegraphics[width=0.7\linewidth]{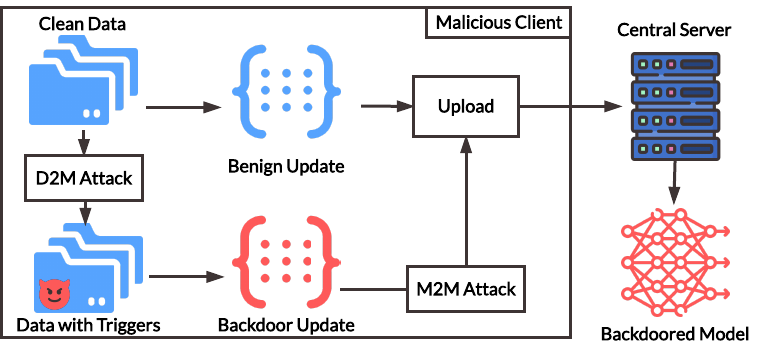}
\caption{A High-level View of Injecting Backdoors with a Composite Attack. The attacker chooses a preferable trigger and tampers local data with the trigger. Local model is also trained on clean data to avoid detection. Most attacks aim at poisoning the global model with only a few clients.}
\label{fig: backdoor}
\end{figure*}

Compared to \ac{D2M} or \ac{M2M} attacks, now that the attacker also has control over client model updates, composite attacks tend to be stealthier and more destructive. A high-level view of such attacks is illustrated in Figure~\ref{fig: backdoor}. We group recent composite attacks based on their most notable features. These attacks may also use techniques proposed in other groups. We show the characteristics of composite attacks in Table~\ref{tab:composite_attack}.

\subsection{Composite Threat Models}

\subsubsection{Update Boosting}

To boost the effectiveness of model updates derived from poisoned data, scaling up malicious updates is a common strategy in early studies on composite attacks \cite{adversarial-lens,how-to-bd}. Given poisoned data with their labels being flipped, authors of \cite{adversarial-lens} propose two types of threat models. The explicit approach is to train client models with the poisoned data, then boost model updates by scaling it up with a predefined coefficient. Although this approach is easy to implement, the boosted updates are statistically different from benign updates, suggesting that secure aggregation rules can easily identify boosted malicious updates. As for the stealthy approach in \cite{adversarial-lens}, the attacker instead trains client models on both the clean and poisoned data. Updates from the poisoned data are boosted as the explicit approach while a regularization term is used to ensure that the differences between current malicious updates and last round's average benign updates are bounded. Instead of boosting only the malicious updates, the model replacement attack proposed in \cite{how-to-bd} seeks to entirely replace the global model with the backdoored model. As the training goes on, benign updates from converging client models tend to cancel each other out. By solving the linear aggregation equation, the attacker can find the solution to scale up malicious updates such that the global model is equal to the model trained with poisoned data, namely the global model is replaced with the one with backdoors.

\subsubsection{Bounded Updates} 

Boosting model updates is an effective way to inject backdoors. However, these updates have distinctive norms compared to benign updates. As mentioned above, boosted updates can be easily filtered out by norm-based aggregation rules. \ac{PGD} proposed in \cite{pgd} aims at bypassing norm-based aggregation by projecting boosted updates onto a small ball around the norm of global model weights. \ac{PGD} can be also seen in later studies \cite{yes-we-can}. On top of the edge case \ac{D2M} attack in \cite{yes-we-can}, the attacker can further cover up their intention by projecting model updates derived from edge case data. Another threat model proposed in \cite{yes-we-can} combines \ac{PGD} with model replacement \cite{how-to-bd} in which the boosted malicious updates is bounded through projection before replacing the global model. Another way to generate bounded updates is proposed in \cite{little-stat}. In stead of projecting malicious updates, they are normalized by the maximum deviation range discussed in the \ac{M2M} attack section.

\subsubsection{Distributed Triggers} 

One common trait of the above composite attacks is that their backdoor triggers are stand-alone, namely the trigger patterns are identical across all clients and tampered samples. Even though there are experiments on concurrently employing multiple triggers \cite{pgd}, these triggers are still independent from each other and they lack the ability to collude. The \ac{DBA} \cite{dba} instead assigns local triggers to multiple clients. Local triggers can be assembled to form a stronger global trigger. The triggers used in \ac{DBA} is similar to the ones used in BadNets \cite{badnets}, which are colored rectangles placed around the corners of images. Malicious updates of \ac{DBA} are scaled up by a coefficient similar to \cite{adversarial-lens}. Another attack with distributed triggers is proposed in \cite{trojan-dba}. Unlike \ac{DBA} whose triggers are predefined, triggers in \cite{trojan-dba} are based on \cite{trojan} with learn-able parameters that generate local trigger patterns. In the trigger generation stage of \cite{trojan-dba}, the attacker first determines the target class. By feeding various samples of the target class to the received global model, the attacker finds the internal neuron that is most sensitive to the target class. This is achieved by comparing the sum of connected weights and the number of activation. The attacker then optimizes trigger pattern parameters such that they maximize the activated value of the most sensitive neuron. In the distributed training stage of \cite{trojan-dba}, each malicious client only trains from the most sensitive neuron's layer to the final output layer. 

\subsubsection{Insidious Tampering} 

More recent composite attacks focus on making malicious updates more insidious and persistent, which is usually achieved by tampering with weights that are unimportant to the clean data. For instance, Neurotoxin \cite{neurotoxin} only updates insignificant parameters to prevent backdoors from being erased by benign updates. Neurotoxin considers parameters with largest gradients to be most used by benign clients, therefore parameters with with smaller gradients are less accessed by benign clients. The attacker can only optimize less important parameters to achieve their backdoor objectives. Neurotoxin is recently enhanced by authors of \cite{rl-neurotoxin} who employ \ac{RL} to find better hyperparameters for the attack. Rare word embedding attack proposed in \cite{rare-word-embedding} shares a similar idea with Neurotoxin in the sense that it manipulates word embeddings of rare words as they are not likely to be updated by benign clients. The effectiveness of the rare word embedding attack can be further amplified by the gradient ensembling method \cite{rare-word-embedding}. The attacker intentionally stores the global models from multiple rounds, then gradients of backdoor word embeddings are computed for all these models. The exponential moving average of these gradients is used to update backdoor embeddings in the current round. \ac{F3BA} is a recent threat model that falls into the category of insidious tampering. Intuitively, \ac{F3BA} tries to flip the signs of lease important weights such that they are most sensitive to trigger patterns. The importance of a weight is measured by the product of its gradient and weight value. \ac{F3BA} only modifies least important weights found by this metric, and empirically $1\%$ of weights are enough to degrade model performance. Sign-flipping of \ac{F3BA} is conducted between consecutive layers. In the first layer, the attacker reshapes the trigger patterns such that it aligns with the convolution kernel. Signs of least important weights of this kernel are flipped if they are different from the signs of the aligned trigger pixels. In subsequent layers, the attacker respectively feeds the model with clean and poisoned data, records their activation differences, and flips signs of the chosen weights such that the activation differences are maximized. When sign-flipping is completed, the model is fine-tuned to associate flipped weights with the labels of poisoned data. The model's local updates will also be more similar to benign updates after fine-tuning. Like \cite{trojan-dba}, trigger patterns is also learn-able. \ac{F3BA} learns the trigger pattern's pixel values by maximizing the clean-poisoned activation difference of the first layer.

\subsubsection{Update Approximation} 

Composite attacks introduced so far directly optimize model weights on the backdoor classification task. There are also attacks seeking to optimize niche objectives. These objectives are often intractable (\emph{e.g.} estimating future updates of other clients), thus the attacker needs to find proper approximations to implement practical solutions. If an omniscient attacker knows all future updates of a \ac{FL} system, the optimal way of injecting backdoors is differentiating through the computation graph of all future updates \emph{w.r.t} the weights of the attacker's model. This is the intuition behind \cite{two-heads} and the authors propose a method to approximate updates in the near future. The attack in \cite{two-heads} requires the attacker to control a subset of client models. The attacker uses these models to simulate future updates by running FedAvg. Throughout the simulation, only clean data sampled from the malicious client is used. In the first round of the simulation, all models are fed with data. The malicious models are left out in the following rounds, which is simulating the scenario in which the malicious client is not chosen by the central server. Once future updates are approximated, client model weights are optimized through the classification losses on both clean and poisoned data similar to \cite{adversarial-lens}. \ac{APA} \cite{apa} is another method that indirectly optimizes model weights for the backdoor task. The objective of \ac{APA} is to clandestinely poison model weights while maintaining a good test performance. As soon as the model is fed with trigger data, its performance drastically drops, leaving the system administrator with minimum time to respond to the attack. \ac{APA} learns two functions: an accumulative function and a poisoning function. The accumulative function is used to manipulate model updates such that the model is more sensitive to trigger gradients. The poisoning function is used to transform benign gradients from validation data into malicious gradients, leading to performance degradation. Intuitively, degrading model performance can be viewed as maximizing the validation loss. By taking the first order Taylor polynomial of the validation loss, the maximization problem is transformed into minimizing the first order gradient \emph{w.r.t} the accumulative and poisoning functions. The authors of \ac{APA} further simplify the minimization problem with its first order approximation. The final optimization objective then becomes simultaneously aligning the directions of poisoned gradients with benign gradients as well as the second order gradients of the validation loss. All gradients from \ac{APA} are all projected through \ac{PGD} \cite{pgd} to enhance stealth. While it is not mandatory to use trigger patterns with \ac{APA}, the authors demonstrate that explicit triggers makes \ac{APA} more potent.


\subsection{Defense Against Composite Attack}

In this section, we introduce defenses that are specifically designed to counter \ac{D2M}+\ac{M2M} composite attacks. Since this type of attack also manipulates model weights or updates, defenses against \ac{M2M} attacks such as Krum \cite{krum} or Bulyan \cite{buylan} are also evaluated in many existing studies on defense against composite attacks. Depending on the subjects being processed by the defense strategy, we divide defenses again composite attacks into update cleansing and model cleansing.

\subsubsection{Update Cleansing} 

Defenses based on update cleansing filter out uploads or mitigate influence from malicious clients by examining model updates. Robust-LR \cite{robust-lr} is an update cleansing defense built on the heuristics that directions of malicious updates are different from benign ones. The authors of Robust-LR take a majority voting over model updates. The voting computes the sum of signs of model updates on each dimension. If the sum is below a pre-defined threshold, meaning that malicious clients participate in the current round of update, the learning rate on that dimension is multiplied by $-1$ to apply gradient ascent to suspicious updates.

Training models with \ac{DP} has been mathematically proven as an effective way of defending against backdoor injections \cite{dp-math, pgd}. This approach is first introduced to \ac{FL} by authors of DP-FedAvg \cite{fl-dp}. Compared to the vanilla FedAvg shown in Algorithm~\ref{alg:fedavg}, DP-FedAvg requires the central server to bound client updates first. Client updates are clipped by comparing its $L2$-norm against a given parameter, which could be an overall parameter for all model weights or a set of layer-wise clipping parameter. When the global model is updated by taking in bounded client updates, noise from a zero-mean Gaussian is also added. 

\subsubsection{Model Cleansing} 

A pruning based method is proposed in \cite{backdoor-pruning}. This approach asks clients to rank the average activation values of the last layer of their models. The central server prunes neurons in the descending order based on the aggregated rankings of neurons. Knowledge distillation is also considered as a defense against composite backdoor attacks \cite{backdoor-distill, f3ba}. By aligning the attention maps of the teacher model and the student model, \ac{NAD} \cite{backdoor-distill} manages to erase backdoors injected in the model. The distillation process of \cite{backdoor-distill} assumes that clean data is available to the defender. This requirement is also inherited by FedRAD \cite{fedrad}, a knowledge distillation based defense for \ac{FL}. FedRAD needs to prepare synthetic data \cite{zero-shot-data} on the central server for model evaluation. Client models are fed with the synthesized data for evaluation, then the central server counts how many times a client's logit obtains the median value for its corresponding class. The median frequencies of client models are normalized and used as global model aggregation coefficients. The distillation process of FedRAD is built on FedDF \cite{feddf}. The central server distills knowledge from client models by minimizing the KL divergence between the global model's predictions and the average prediction of client models. 

Some research considers certified robustness \cite{certified-robust} as the way to defend against composite backdoor attacks. A \ac{ML} model is said to have certified robustness if its predictions are still stable even if the input is perturbed. CRFL \cite{crfl} is a defense designed to counter the model replacement attack. By controlling how the global model parameters update during training, CRFL grants the global model certified robustness under the condition that the backdoor trigger is bounded. Specifically, when the conventional global model aggregation completes, parameters of the global model are first clipped, then Gaussian noise is added to these parameters. At test time, a set of Gaussian noise is sampled from the previous noise distribution and added to the aggregated global model, resulting in a set of noisy global models. A majority voting is conducted among these noisy models to decide the classification results of test samples. Another defense with certified robustness is proposed in \cite{another-cert}. This method achieves certified robustness through the majority voting among a number of concurrently trained global models. Given $n$ clients, the defense in \cite{another-cert} trains $\binom nk$ global models, where $k$ is the number clients chosen without replacement for each model. Although the authors of \cite{another-cert} applies Monte Carlo approximation to speed up the defense, it still needs to train hundreds of global models, making this method more computationally expensive than other defenses.

The idea of majority voting is not exclusive to defenses with certified robustness. Authors of BaFFLe \cite{baffle} rely on diversified client data to validate and provide feedback to the global model. BaFFLe adds an extra stage to conventional \ac{FL} pipeline. When the global model for current global training round is aggregated, it is sent to randomly selected clients to validate if the global model is poisoned. A set of recently accepted global models are also sent to selected clients as reference. The validation process s of BaFFLe requires these clients to test global models with their local data. In particular, each client computes the misclassification rate for samples of a specific class, the client also computes the rate of other classes’ samples being misclassified as the examined class. For benign models, the gap between these two rates are relatively stable during training. However, drastic changes can happen for backdoored models. If the misclassification gap of the newly aggregated global model deviates too much from the average gap of past models, the client votes the global model as malicious. Finally, based on the result of the majority voting, the central server decides whether to discard the newly obtained global model.

\subsubsection{Composite Cleansing} 

Like composite attacks that manipulate multiple aspects of \ac{FL} to enhance their capability, recent defenses also examine both model updates and weights to systematically mitigate composite attacks.

Authors of DeepSight \cite{deepsight} propose various metrics to evaluate if the upload from a client is malicious. The central server first computes the pairwise cosine similarities between received updates. Two other metrics, clients’ \ac{DDif} and \ac{NEUP}, are also computed. \ac{DDif} measures the prediction differences between the global and client models. This is achieved by feeding models with random input on the server. Backdoored models are prone to produce larger activation for the trigger class even if the input is merely random noise \cite{central-trojan}, which is a telltale sign for \ac{DDif} to identify compromised models. \ac{NEUP} measures the update magnitude for neurons in the output layer. Local data with similar distributions results in models with similar \ac{NEUP} patterns. Based on the above metrics, DeepSight clusters received client models on the central server with HDBSCAN \cite{hdbscan}. The server also needs to maintain a classifier based on \ac{NEUP} to label client models as either benign or malicious. Depending on the number of models being labeled as malicious, the server determines whether to accept or reject a client model cluster. Models from accepted clusters are deemed as safe for aggregation.

FLAME \cite{flame} is another example of composite defense. Authors of FLAME summarize the pipeline of their approach as clustering, clipping and noising. In the clustering stage, the central server computes \ac{CD}s between model updates. HDBSCAN is subsequently used to filter out malicious models based on the angular differences derived from \ac{CD}s. In the clipping stage, the median of remaining models’ updates is chosen as the bound to clip model updates. In the final noising stage, Gaussian noise is added to the global model weights to further erase injected back doors. 

\section{Conclusion and Future Directions}
\label{sec:con}

\subsection{Conclusion}

In recent years, \ac{FL} has become a transformative paradigm for training \ac{ML} models, especially in decentralized environments where data privacy and security are critical. Our comprehensive review categorized known \ac{FL} attacks according to attack origin and target. It provides a clear structure for understanding the scope and depth of \ac{FL} inherent vulnerabilities:
\newline

\noindent\textbf{\ac{D2M} Attacks:} These attacks (\emph{e.g.}, label-flipping) manipulate data to corrupt the global model. Since \ac{FL} often relies on data from numerous potentially untrusted sources, it is highly vulnerable to such threats.
\newline

\noindent\textbf{\ac{M2M} Attacks:} This type of attack tampers with model updates, thereby disrupting the learning process. For example, Byzantine attacks involve sending malformed or misleading model updates, indicating that one or more malicious clients have the potential to degrade the performance of the global model. Such attacks emphasize the importance of a robust aggregation approach in a federated environment.
\newline

\noindent\textbf{\ac{M2D} Attacks:} Focus on exploiting vulnerabilities that arise when models interact with data, such as gradient leakage, where an attacker can infer private data from gradient updates. Gradient leakage is a prime example where malicious entities exploit the shared model updates to infer sensitive information about the training data, emphasizing on the need for defense strategies that mask or generalize gradients.
\newline

\noindent\textbf{Composite Attacks:} These attacks are more sophisticated in nature and often combine multiple attack methods or vectors to enhance their impact. Backdoor injection is a classic example, where an attacker subtly introduces a backdoor during training and then exploits it during reasoning.

A summarization of defense techniques toward different types of attacks is provided in Table~\ref{tab:summ_defense}

\begin{table*}[!ht]
\setlength{\tabcolsep}{1pt}
\centering
\caption{Summarization of defense techniques toward different types of attacks}
\begin{tabular}{c | c | c | c}
\hline
Defense Method & Defense Strategy & Type of Attack & Attack Strategy  \\
\hline
\makecell[c]{Fung et al.\cite{sybil-attack} (FoolsGold)\\ Tolpegin et al.\cite{detailed-label-flip} \\ Cao et al.\cite{understanding-label-flipping} (Sniper) \\ Ma et al.\cite{frl-d2m}}& \makecell[c]{Dynamic learning rate \\ Cluster for PCA \\ Clique from Euclidean distance \\ Rewards based aggregation} & \ac{D2M} & \makecell[c]{Label Attack\\Sample Attack}  \\
\hline
\makecell[c]{Chen et al.\cite{geomed} (GeoMed) \\ Pillutla et al.\cite{improved-gm} (RFA) \\ Xie et al.\cite{generalized-bSGD} (MarMed) \\ Xie et al.\cite{generalized-bSGD} (MeaMed) \\ Yin et al.\cite{coordinate-median} (TrimMean) \\ Blanchard et al.\cite{krum} (Krum) \\ El Mhamdi et al.\cite{buylan} (Bulyan) \\ Wang et al.\cite{elite} (ELITE) \\ Tekgul et al.\cite{waffle} (WAFFLE) \\ Li et al.\cite{fed-ipr} (FedIPR) \\ Lin et al.\cite{gaussian-rider} \\ Zong et al.\cite{dagmm} (DAGMM)} & \makecell[c]{Geometric median \\ Weiszfeld-smoothed geometric median \\ Dimension-wise median \\ Mean-around median \\ Dimension-wise trimmed mean \\ Euclidean distance \\ Euclidean distance\\ Gradient information gain \\ The server embeds watermarks \\ Generate secret watermarks on client \\ Auto-encoder \\ Gaussian mixture network} & \ac{M2M} & \makecell[c]{Priori Attack\\Posteriori Attack} \\
\hline
\makecell[c]{Zhu et al.\cite{zhu2019deep} \\ Chamikara et al.\cite{chamikara2021privacy} \\ Wei et al.\cite{wei2021gradient} \\ Scheliga et al.\cite{scheliga2022precode} (PRECODE) \\ Ren et al.\cite{ren2023gradient} (FedKL)} & \makecell[c]{Adding noise to gradients \\ Perturbing data \\ DP on data \\ Transform feature representation \\ Hide the input from gradient} & \ac{M2D} & \makecell[c]{Attribute Inference\\Membership Identification\\Image Recovery} \\
\hline
\makecell[c]{Ozdayi et al.\cite{robust-lr} (Robust-LR) \\ McMahan et al.\cite{fl-dp} (DP-FedAvg) \\ Wu et al.\cite{backdoor-pruning} \\ Sturluson et al.\cite{fedrad} (FedRAD) \\ Xie et al.\cite{crfl} (CRFL) \\ Cao et al.\cite{another-cert} \\ Andreina et al.\cite{baffle} (BaFFLe) \\ Rieger et al.\cite{deepsight} (DeepSight) \\  Nguyen et al.\cite{flame} (FLAME)} & \makecell[c]{Update cleansing \\ DP \\ Model pruning \\ Knowledge distillation \\ Certified robustness from updates \\ Certified robustness  \\Validation on diversified client data \\ Various metrics \\ Clustering, clipping and noising} & Composite & \makecell[c]{Updates Attack\\ Distributed Triggers \\ Insidious Tampering} \\
\hline
\end{tabular}
\label{tab:summ_defense}
\end{table*}
\subsection{Future Directions}

As \ac{FL} continues to evolve, the sophistication of potential attacks will continue to increase. By reviewing the recent advancements in this domain, we identify several promising research directions that include:
\newline

\noindent\textbf{Robust Aggregation Mechanisms:} The aggregation process in \ac{FL} is a key link where local model updates from different participants are combined to update the global model. Given its central role, the aggregation step becomes a vulnerable point, especially to malicious interference. For example, a single participant with malicious intentions may submit misleading updates with the intention of degrading the performance of the global model. This adverse activity is of particular concern in \ac{M2M} attacks, of which the Byzantine attack is a prime example. In a Byzantine attack, an adversary sends arbitrary or strategically designed updates to a server with the intent of disrupting the aggregated model. Addressing these vulnerabilities requires re-evaluating and redesigning the traditional aggregation mechanisms used in \ac{FL}. By delving into the development of more resilient aggregation strategies, methods can can be designed to identify, isolate, or reduce the impact of these malicious updates. These advanced aggregation techniques, based on robust statistical measures, consensus algorithms and even outlier detection methods, can ensure that the integrity of the global model remains intact in the presence of hostile participants.
\newline

\noindent\textbf{Gradient Sparse Attack:}
In terms of \ac{M2D} attack methods, it is worth noting that the gradients exchanged between the server and the client often contain a large amount of redundant details~\cite{yin2021see}, and this redundancy may play a negative role in the effectiveness of the attack. If an attacker can filter out valuable gradients, the efficiency of the attack can be dramatically improved, especially in large-scale model training. This gradient sparse process eliminates irrelevant and noisy data, thus potentially improving the accuracy of the attack.
\newline

\noindent\textbf{Automatic Attack Detection:} As the complexity and scale of \ac{FL} environments continues to grow, automated safety measures become critical. Meta-learning~\cite{finn2017model,snell2017prototypical,lee2019meta,cao2022relational}, often referred to as ``learning to learn'', offers a promising avenue to address this challenge. By employing meta-learning techniques, systems can be trained to leverage prior knowledge about different types of attacks to quickly adapt to new, unforeseen threats. In addition, anomaly detection algorithms help identify outliers or unusual patterns in traditional datasets that can be fine-tuned for federated environments. These algorithms can monitor incoming model updates from different clients or nodes and flag any updates that deviate from the expected pattern to indicate potential malicious activity. Such an automated system not only identifies threats, but also combines with defense mechanisms to immediately counteract or eliminate suspicious activity, ensuring a smoother and safer \ac{FL} process.
\newline

\noindent\textbf{Holistic Defense Strategies:} In the rapidly evolving \ac{FL} environment, the need for holistic defense strategies is becoming increasingly prominent. These strategies advocate the development and implementation of defense mechanisms that are inherently versatile and capable of responding to multiple attack vectors simultaneously. A holistic approach would integrate various protection measures to create a more resilient and adaptive security framework, rather than a solo approach that develops defenses against specific threats. This multi-pronged defense system not only ensures broader security coverage, but also minimizes potential vulnerabilities and overlaps. As adversarial tactics become increasingly complex, utilizing an integrated solution that anticipates and responds to a wide range of threats will be key to protecting the \ac{FL} ecosystem.
\newline

\noindent\textbf{Domain-specific Attacks and Defenses}
Although we have witnessed nascent studies on exploiting the vulnerabilities in Federated Recommendation System and Federated \ac{RL}, few defenses are proposed to defend against such threats. Furthermore, a majority of the current research tends to focus on image classification as the principal learning task for both attacks and defenses. This observation underscores a pressing need and opportunity to delve deeper into domain-specific threat models and tailored defense strategies for federated learning. Investigating this avenue not only holds promise for enhancing security but also ensures the more comprehensive protection of diverse applications within \ac{FL}.
\newline

\noindent\textbf{Interdisciplinary Approaches:} Harnessing the wealth of insights from different fields is particularly instructive for enhancing \ac{FL} systems. For example, frameworks and theories from disciplines such as game theory and behavioral science can help to understand the motivations and behaviors of participants in a \ac{FL} environment. By understanding these motivations, tailored incentive structures or deterrence mechanisms can be designed to encourage positive contributions and discourage malicious or negligent behaviors in \ac{FL} ecosystems. In addition, the fields of cryptography and cyber-security are constantly evolving, offering a plethora of innovative techniques and protocols. By integrating these advances into \ac{FL}, we can strengthen systems against identified vulnerabilities and ensure not only the privacy and integrity of data, but also the trustworthiness of the learning process. As the stakes for \ac{FL} grow, especially in critical areas of application, the convergence of these areas is critical to creating a robust, secure and collaborative learning environment.




\bibliographystyle{elsarticle-num}
\bibliography{citations}

\begin{thebibliography}{100}
\bibitem{weiszfeld}
E.~Weiszfeld, F.~Plastria, On the point for which the sum of the distances to n
  given points is minimum, Ann Oper Res (2009).
\end{thebibliography}

\end{document}